# Improvements in Sub-optimal Solving of the $(N^2 - 1)$-Puzzle via Joint Relocation of Pebbles and its Applications to Rule-based Cooperative Path-Finding

**Pavel Surynek**[1] and **Petr Michalík**[2]

[1]*Charles University*
*Faculty of Mathematics and Physics*
*Department of Theoretical Computer Science and Mathematical Logic*
*Malostranské náměstí 25, Praha, 118 00, Czech Republic*

[2]*Accenture Central Europe B.V. Consulting*
*Jiráskovo náměstí 1981/6, Praha, 120 00, Czech Republic*

pavel.surynek@mff.cuni.cz, petr.michalik@accenture.com

*Abstract.* The problem of solving $(n^2 - 1)$-puzzle and cooperative path-finding (CPF) sub-optimally by rule based algorithms is addressed in this manuscript. The task in the puzzle is to rearrange $n^2 - 1$ pebbles on the square grid of the size of $n \times n$ using one vacant position to a desired goal configuration. An improvement to the existent polynomial-time algorithm is proposed and experimentally analyzed. The improved algorithm is trying to move pebbles in a more efficient way than the original algorithm by grouping them into so-called snakes and moving them jointly within the snake. An experimental evaluation showed that the algorithm using snakes produces solutions that are 8% to 9% shorter than solutions generated by the original algorithm.

The snake-based relocation has been also integrated into rule-based algorithms for solving the CPF problem sub-optimally, which is a closely related task. The task in CPF is to relocate a group of abstract robots that move over an undirected graph to given goal vertices. Robots can move to unoccupied neighboring vertices and at most one robot can be placed in each vertex. The $(n^2 - 1)$-puzzle is a special case of CPF where the underlying graph is represented by a 4-connected grid and there is only one vacant vertex. Two major rule-based algorithms for CPF were included in our study – *BIBOX* and *PUSH-and-SWAP* (*PUSH-and-ROTATE*). Improvements gained by using snakes in the *BIBOX* algorithm were stable around 30% in $(n^2 - 1)$-puzzle solving and up to 50% in CPFs over bi-connected graphs with various ear decompositions and multiple vacant vertices. In the case of the *PUSH-and-SWAP* algorithm the improvement achieved by snakes was around 5% to 8%. However, the improvement was unstable and hardly predictable in the case of *PUSH-and-SWAP*.

## 1. Introduction and Motivation

The $(n^2 - 1)$-puzzle [9, 10, 12, 13] represents one of the best-known examples of a so-called *cooperative path-finding* (CPF) [23, 33, 38, 42] problem. It is important both practically and theoretically. From the theoretical point of view it is interesting for the hardness of its optimization variant which is known to be $NP$-hard [16, 17].

Practically it is important since many real-life relocation problems can be solved by techniques developed for the $(n^2 - 1)$-puzzle. Those include *path planning for multiple* *robots* [1, 8, 19, 20, 26, 29, 25, 39, 42], *trajectory planning* [5, 7], *rearranging* of shipping containers in warehouses, or *coordination* of vehicles in dense traffic. Moreover, the reasoning about relocation/coordination tasks should not be limited to physical entities only. Many tasks such as planning of *data transfer*, *commodity transportation*, and *motion planning* of units in *computer-generated imagery* can be tackled using techniques originally developed for the $(n^2 - 1)$-puzzle.

In this manuscript, we concentrate on solving the $(n^2 - 1)$-puzzle sub-optimally, that is, by fast polynomial-time algorithms. We are trying to improve the basic incremental placing of pebbles as it is done by the existent state-of-the-art on-line solving algorithm of *Parberry* [13] by moving them jointly in groups called *snakes*. Moving pebbles jointly in snakes is supposed to be more efficient in terms of the total number of moves than moving them individually as it was originally proposed [13]. An improved algorithm exploiting snake-based movements is presented.

We utilized experiences gained during making snake-based improvements to *Parberry's* algorithm in solving of CPFs sub-optimally by rule-based algorithms. We took existing rule-based cooperative path finding algorithms and conducted a study how they can be improved via joint robot movements in snakes. There are two rule-based algorithms for CPF – *BIBOX* [26, 33] and *PUSH-and-SWAP* (*PUSH-and-ROTATE*) [12, 42, 43] similar to the *Parberry's* algorithm. Both algorithms operate in a similar way to the algorithm of *Parberry*; that is, they also place robots one by one to their goal positions as it is done by *Parberry's* algorithm with pebbles. Hence, the snake-based reasoning for joint movements of pebbles can be applied with certain effort within these algorithms as well. The *BIBOX* algorithm is originally developed for CPFs over bi-connected graphs with at least two vacant vertices while *PUSH-and-SWAP* is more general – it can be applied to CPF over arbitrary graph with at least two vacant vertices. Both rule-based algorithms are thus applicable the puzzle where one more blank is added – that is to $(n^2 - 2)$-puzzle – which allows us to make competitive comparison of all the algorithms.

The snake-based reasoning is focused on improving solutions generated by selected algorithms in terms of *the number of moves*. In this context, we need to mention a great progress that has been done in solving the CPF problem as well as the $(n^2 - 1)$-puzzle optimally with respect to various objectives such as the *total number of moves* [21], *parallel makespan* [34], and derivations of these. There exist great variety of search-based optimal algorithms such as *ID+OD* (*Independence Detection + Operator Decomposition*) [25], *ICTS (Increasing Cost Tree Search)* [21], and *CBS* (*Conflict-based Search*) [22] to name few. Reductions of optimal CPF solving to other formalisms such as *SAT* (*Propositional Satisfiability*) [34] or *ASP* (*Answer Set Programming*) [6] allow using external solvers for a given formalism. Although mentioned optimal approaches are still improved to be able to solve larger instances optimally they are far in term of size from the instances that can be solved by studied sub-optimal approaches. While either the optimal techniques need few robots only or sparse environments, the sub-optimal approaches can tackle large densely occupied instances.

An extensive competitive experimental evaluation was done to evaluate qualities of snake-based improvements in solving $(n^2 - 1)$-puzzle as well as in solving of CPFs over bi-connected graphs. All the algorithms were tested against their variant with snake-based reasoning on a number of benchmark instances.

The manuscript is organized as follows. The problem of $(n^2 - 1)$-puzzle is formally introduced in Section 2. An overview of existent solving algorithm and other related solving approaches is given in Section 3. The main part of the paper is constituted by Section 4, Section 5, and Section where the *snake movement* is introduced into the Parberry's algorithm, into *BIBOX*, and into *PUSH-and-SWAP*. Although these algorithms share certain similarities, they are also fundamentally different. Hence, the snake-based reasoning needed to be adapted for each algorithm substantially. An extensive experimental evaluation is finally given in Section 6.

## 2. Problem Statement

The $(n^2 - 1)$-puzzle consists of a set of pebbles that are moved over a square grid of size $n{\times}n$ [1, 13, 16, 17, 44]. There is exactly one position vacant on the grid and others are occupied by exactly one pebble. A pebble can be moved to the adjacent vacant position. The task is to rearrange pebbles on the grid into a desired goal state.

### 2.1. *Formal Definition*

Sets of pebbles will be denoted as $\Omega_n$ for $n \in \mathbb{N}$. It holds that $|\Omega_n| = n^2 - 1$ for every $n \in \mathbb{N}$. It is supposed that pebbles from a set $\Omega_n$ are arranged on a square grid of the size $n{\times}n$ where each pebble is placed into one of the cells of the grid. There is at most one pebble in each cell of the grid; one cell on the grid remains always vacant (Figure 1).

**Definition 1 *(configuration in a grid).*** An *configuration* of a set of pebbles $\Omega_n$ in a square grid of the size $n{\times}n$ with $n \in \mathbb{N}$ is fully described using two functions $x_n: \Omega_n \longrightarrow \mathbb{N}$ and $y_n: \Omega_n \longrightarrow \mathbb{N}$ that satisfy the following *puzzle conditions*:

(i) $x_n(p) \in \{1,2,\dots,n\}$ and $y_n(p) \in \{1,2,\dots,n\}$ $\forall p \in \Omega_n$ (1)

(ii) $|\{p \in \Omega_n | (x_n(p), y_n(p)) = (i,j)\}| \leq 1$ for $\forall i,j \in \{1,2,\dots,n\}$ (2)
(every cell of the grid is occupied by at most one pebble)

(iii) $\exists i,j \in \{1,2,\dots,n\}$ such that $\forall p \in \Omega_n$ $(x_n(p), y_n(p)) \neq (i,j)$ (3)
(there exists a cell in the grid that remains vacant).

For convenience, we will also use some kind of an inverse to $x_n$ and $y_n$ which will be called an *occupancy* function and denoted as $\sigma_n: \{1,2,\dots,n\}{\times}\{1,2,\dots,n\} \longrightarrow \Omega_n \cup \{\bot\}$. It holds that $\sigma_n(i,j) = p$ if and only if $p \in \Omega_n$ and $x_n(p) = i$ and $y_n(p) = j$ or $\sigma_n(i,j) = \bot$ if no such pebble $p$ exists (that is, if the cell $(i,j)$ is vacant). □

The configuration of pebbles in the grid can be changed through moves. An allowed *move* is to shift a pebble horizontally or vertically from its original cell to the adjacent vacant cell. Formally, the notion of move is described in the following definition. Four types of moves are distinguished here: *left*, *right*, *up*, and *down* – only left move is defined formally; *right*, *up*, and *down* moves are analogous.

**Definition 2 *(left move).*** A *left* move with pebble $p \in \Omega_n$ can be done if $x_n(p) > 1$ and $\sigma_n(x_n(p) - 1, y_n(p)) = \perp$; it holds for the resulting configuration after the move described by $x'_n$ and $y'_n$ that $x'_n(q) = x_n(q)$ and $y'_n(q) = y_n(q)$ $\forall q \in \Omega_n$ such that $q \neq p$ and $x'_n(p) = x_n(p) - 1$ and $y'_n(p) = y_n(p)$. □

We are now able to define the $(n^2 - 1)$-puzzle using the formal constructs we have just introduced. The task is to transform a given initial configuration of pebbles in the grid to a given goal one using a sequence of allowed moves.

**Definition 3 *(($\boldsymbol{n^2 - 1}$)-puzzle).*** An *instance of the* $(n^2 - 1)$*-puzzle* is a tuple $(n, \Omega_n, x_n^0, y_n^0, x_n^+, y_n^+)$ where $n \in \mathbb{N}$ is the size of the instance, $\Omega_n$ is a set of pebbles, $x_n^0$ and $y_n^0$ is a pair of functions that describes the *initial configuration* of pebbles in the grid, and $x_n^+$ and $y_n^+$ is a pair of functions that describes the *goal configuration* of pebbles. The task is to find a sequence of allowed moves that transforms the initial configuration into the goal one. Such sequence of moves will be called a *solution* to the instance. □

Again it is supposed that the occupancy function is available with respect to the initial configuration $x_n^0, y_n^0$ and the goal configuration $x_n^+, y_n^+$; that is, we are provided with occupancy functions $\sigma_n^0$ and $\sigma_n^+$. To avoid special cases it will be also supposed that $\sigma_n^+(n, n) = \perp$; that is, the vacant position is finally in the right bottom corner.

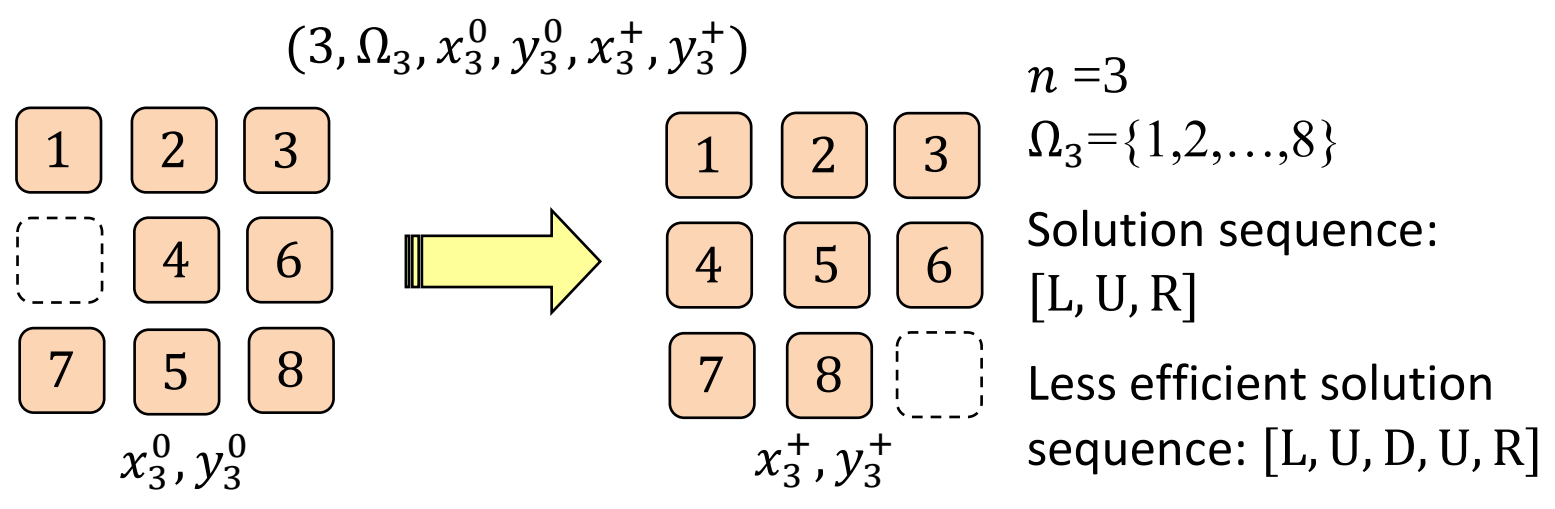

**Figure 1.** *An illustration of the* $(n^2 - 1)$*-puzzle.* The initial and the goal configuration of pebbles on the square grid of size 3×3 are shown. Two solutions of the instance are shown as well.

### 2.2. ***Complexity and Variants of the Problem***

It is known that the decision variant of the $(n^2-1)$-puzzle (that is, the yes/no question whether there exists a solution to the given instance) is in $P$ [1, 13, 44]. It can be checked by using simple parity criterion. Using techniques for rearranging pebbles over graphs [1] a solution of length $\mathcal{O}(n^6)$ can be constructed in the worst-case time of $\mathcal{O}(n^6)$ if there exists any. An approach dedicated exclusively to the $(n^2-1)$-puzzle is able to generate a solution of length $\mathcal{O}(n^3)$ in the worst-case time of $\mathcal{O}(n^3)$ if there exists any [13].

If a requirement on the length of the solution is added, the problem becomes harder. It is known that the decision problem of whether there exists a solution to a given $(n^2-1)$-puzzle of at most the given length is $NP$-complete [17].

## 3. The Original Solving Algorithm and Related Works

A special sub-optimal solving algorithm dedicated for the $(n^2-1)$-puzzle has been proposed by *Parberry* in [13]. As our new solving algorithm is based on the framework of the original one, we need to recall it at least briefly in this section.

### 3.1. ***Algorithm of Parberry***

The algorithm of *Parberry* [13] sequentially places pebbles into rows and columns. More precisely, pebbles are placed sequentially into the first row and then into the first column, which reduces the instance to that of the same type but smaller – that is, we obtain an instance of the $((n-1)^2-1)$-puzzle.

This process of pebble placement is repeated until an 8-puzzle on the grid of size 3×3 is obtained. The final case of the 8-puzzle is then solved optimally by the A* algorithm [18].

The main loop of the algorithm is shown in pseudo-code as Algorithm 1. The algorithm uses two high-level functions *Place-Pebble*, which conducts placement of a pebble to a given position, and *Solve-8-Puzzle*, which finalizes the solution by solving the residual 8-puzzle.

The placement of pebbles implemented within the function *Place-Pebble* will be discussed in more details later in the context of our improvement. Nevertheless, it is done quite naturally by moving a pebble first *diagonally* towards the goal position if necessary and then *horizontally* or *vertically*. To be able to conduct diagonal, horizontal and vertical movement a vacant position needs to be moved together with the pebble being placed. Actually, the vacant position is moving around the pebble always to the front in the direction of the intended move. After having vacant position in the front, the pebble is moved forward. It is necessary to avoid already placed pebbles when placing a new one.

**Algorithm 1.** *The original algorithm of Parberry for solving the* $(n^2-1)$*-puzzle* [13]. The main loop of the algorithm is shown. Detailed description of placement of individual pebbles is not shown here –it will be discussed in the context of new approach for pebble placement.

**procedure** *Solve-N^2-1-Puzzle*$(n, \Omega_n, x_n^0, y_n^0, x_n^+, y_n^+)$

/* A procedure that produces a sequence of moves that solves the given $(n^2-1)$-*puzzle*.
Parameters: $n, \Omega_n$ - a size of the puzzle and a set of pebbles,
$x_n^0, y_n^0$ - an initial configuration of pebbles in the grid,
$x_n^+, y_n^+$ - a goal configuration of pebbles in the grid. */

$(x_n, y_n) \leftarrow (x_n^0, y_n^0)$
for $i = 1,2,\ldots,n-3$ do
  for $j = i, i+1, \ldots, n$ do {current row is solved – from the left to the right}
    $p \leftarrow \sigma_n^+(i,j)$
    if $(i,j) \neq (x_n^+(p), y_n^+(p))$ then
      $(x_n, y_n) \leftarrow$ *Place-Pebble*$(x_n, y_n, i, j, p)$
    $\Omega_n \leftarrow \Omega_n \setminus \{p\}$
  for $j = n, n-1, \ldots, i+1$ do {current column is solved – from the bottom to the up}
    $p \leftarrow \sigma_n^+(i,j)$
    if $(i,j) \neq (x_n^+(p), y_n^+(p))$ then
      $(x_n, y_n) \leftarrow$ *Place-Pebble*$(x_n, y_n, i, j, p)$
    $\Omega_n \leftarrow \Omega_n \setminus \{p\}$
$\Omega_3 \leftarrow \Omega_n$; $x_3^0 \leftarrow x_n|_{\Omega_3}$; $y_3^0 \leftarrow y_n|_{\Omega_3}$; $x_3^+ \leftarrow x_n^+|_{\Omega_3}$; $y_3^+ \leftarrow y_n^+|_{\Omega_3}$ {restriction on $\Omega_n$}
*Solve-8-Puzzle*$(\Omega_3, x_3^0, y_3^0, x_3^+, y_3^+)$ {the residual 8-puzzle is solved by A* algorithm}

## 3.2. *Other Related Works*

The $(n^2-1)$-puzzle represents a special variant of a more general problem of *cooperative path-finding* - CPF (also known as *pebble motion problem on a graph*) [10, 11, 19, 20, 23, 29, 44]. The generalization consists in the fact that there is an arbitrary undirected graph representing the environment instead of the regular 4-connected grids as it is in the case of $(n^2-1)$-puzzle. There are also pebbles, in context of CPF called *robots* (or agents), that are placed in vertices of the graph while at least one vertex remains vacant. The allowed state transition is a single move with a robot to a vacant adjacent vertex. The task is expectably to rearrange robots from a given initial configuration to a given goal one.

Although the problem has been already studied [10, 44], new results appeared recently. One of recent works shows solvability of every instance of pebble motion problem consisting of bi-connected graph [36, 40, 41] containing at least two vacant positions [26]. The related solving algorithm called *BIBOX* [26] can produce solution of length at most $\mathcal{O}(|V|^3)$ in the worst-case time of $\mathcal{O}(|V|^3)$ ($V$ is the set of vertices of the input graph). The *BIBOX* algorithm also generates solutions that are significantly shorter than those generated by algorithms from previous works [10, 44].

A generalization of *BIBOX* algorithm called *BIBOX-θ* is described in [29]. It does not need the second vacant position and again can solve instances on bi-connected graphs (no-tice that the grid of the $(n^2 - 1)$-puzzle is a bi-connected graph; hence *BIBOX-θ* is applicable to it). Theoretically, it generates solutions of the worst-case length of $\mathcal{O}(|V|^4)$; however, practically solutions are much shorter.

A more general algorithm called *PUSH-and-SWAP* has been published in [11] – it shows that for every solvable instance on an arbitrary graph containing at least two vacant positions a solution of length $\mathcal{O}(|V|^3)$ can be generated. The algorithm omits few cases that make it incomplete; however a corrected version of *PUSH-and-SWAP* called *PUSH-and-ROTATE* has been appeared in [42, 43].

In all the above results the solution length is sub-optimal and the worst-case time complexity is guaranteed (it is polynomial). A progress has been also made in optimal solving of the pebble motion problem. A new technique that can optimally solve a special case consisting of a grid with obstacles and relatively small number of pebbles is described in [25] as *ID+OD* algorithm *(Independence Detection + Operator Decomposition)*. It is based on an informed search with powerful heuristics, which however does not guarantee time necessary to produce a solution (the time may be exponential in the size of the instance) as in the case of sub-optimal methods studied in this work. A more recent progress in search-based techniques for optimal CPF solving is represented by [3, 4, 21, 22].

Special cases of the problem with large graphs and relatively sparsely arranged pebbles are studied in [38, 39]. These new techniques are focused on applications in *computer games*. The complexity as well as the solution quality is guaranteed by these techniques. Another specialized technique for relatively large graphs and small number of pebbles has been developed within [19, 20]. The graph representing the environment is decomposed into subgraph patterns, which are subsequently used for more efficient solving by search.

## 4. A New Solving Approach Based on ‘Snakes’

In this section, we are about to define a new concept of a so-called *snake*. Informally, a snake is a sequence of pebbles that consecutively neighbors with a pebble that proceeds. As we will show, moving and placing a snake as a whole is much more efficient than moving and placing individual pebbles it consists of.

Recall that original algorithm for solving the puzzle [13] places pebbles individually into currently solved row or column. This may be inefficient if two or more pebbles that need to be placed are grouped together in some location distant from their goal location. In such a case, it is necessary that the vacant position is moved together with the pebble being placed and then it is moved back to the distant location to allow movement of the next pebble. If we manage all the pebbles forming the group to move from their distant location to their goal positions jointly, multiple movements of the vacant position between the distant location and goal positions may be eliminated.

### 4.1. *Formal Definition of a 'Snake'*

Consider a situation shown in Figure 2 where pebbles 1 and 2 are grouped together in a location distant from their goal positions. The original algorithm consumes $16n - 20$ moves to place both pebbles successfully to their goal positions. If pebbles are moved not one by one but jointly as it is shown in Figure 3, much less movements are necessary. Grouping pebbles can save up to $4 \cdot n$ moves.

This is the basic idea behind the concept of snake. Let us start with definition of a *metric* on the grid of the puzzle. Then the definition of the snake will follow.

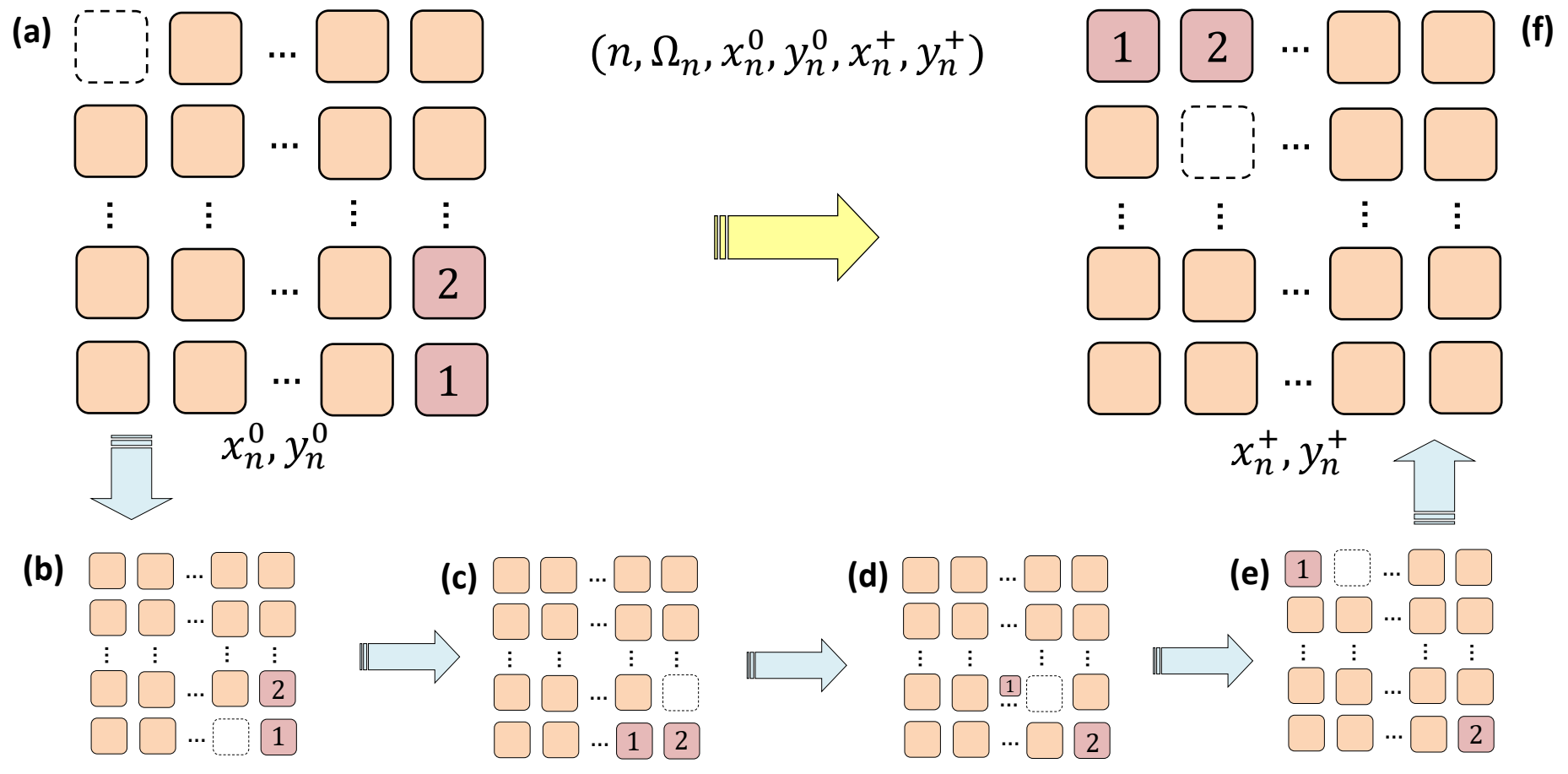

**Figure 2.** *A setup of the* $(n^2 - 1)$*-puzzle where the original algorithm* [13] *is inefficient.* Pebbles 1 and 2 need to be moved from the bottom right corner (a) to the upper left corner (f). First, pebble 1 is moved diagonally to its goal position (b, c, d, and e). After pebble 1 is successfully placed, vacant position is moved towards pebble 2 and it starts to move in the same way as pebble 1 to its goal position. The whole process of rearranging consumes $16n - 20$ moves.

**Definition 4** ***(Manhattan distance).*** A *Manhattan distance* for the $(n^2 - 1)$-puzzle $\mu_n$: $\{1,2,\dots,n\}^2 \times \{1,2,\dots,n\}^2 \longrightarrow \{0,1,\dots,2n-1\}$ is a metric on the square grid such that $\mu_n((x_1, y_1); (x_2, y_2)) = |x_1 - x_2| + |y_1 - y_2|$. □

The input parameters of the Manhattan distance $\mu$ are coordinates of two positions. Having a metric on the grid of the puzzle, we are able to define neighborhood of a pebble. A snake will be then defined using the notion of neighborhood as a sequence of pebbles that consecutively lies in neighborhood of a pebble that proceeds.

**Definition 5** ***(Manhattan neighborhood).*** A *Manhattan neighborhood* of a pebble $p$ denoted as $\nu(p)$ is a set of those pebbles that are located directly left, right, above and below to $p$ with respect to the configuration on the grid. That is, $\nu(p) = \{q \in \Omega_n | \mu_n((x_n(p), y_n(p)); (x_n(q), y_n(q))) = 1\}$. □

**Definition 6 *(Snake).*** A *snake* $s$ of size $k$ is a sequence of pebbles $s = [s_1, s_2, \dots, s_k]$ such that $\forall i \in \{1,2,\dots,k\}$ $s_i \in \Omega_n$ and $\forall j \in \{2,3,\dots,k\}$ $s_j \in \nu(s_{j-1})$. Pebble $s_1$ is called a *head* of the snake; pebble $s_k$ is called a *tail* of the snake. □

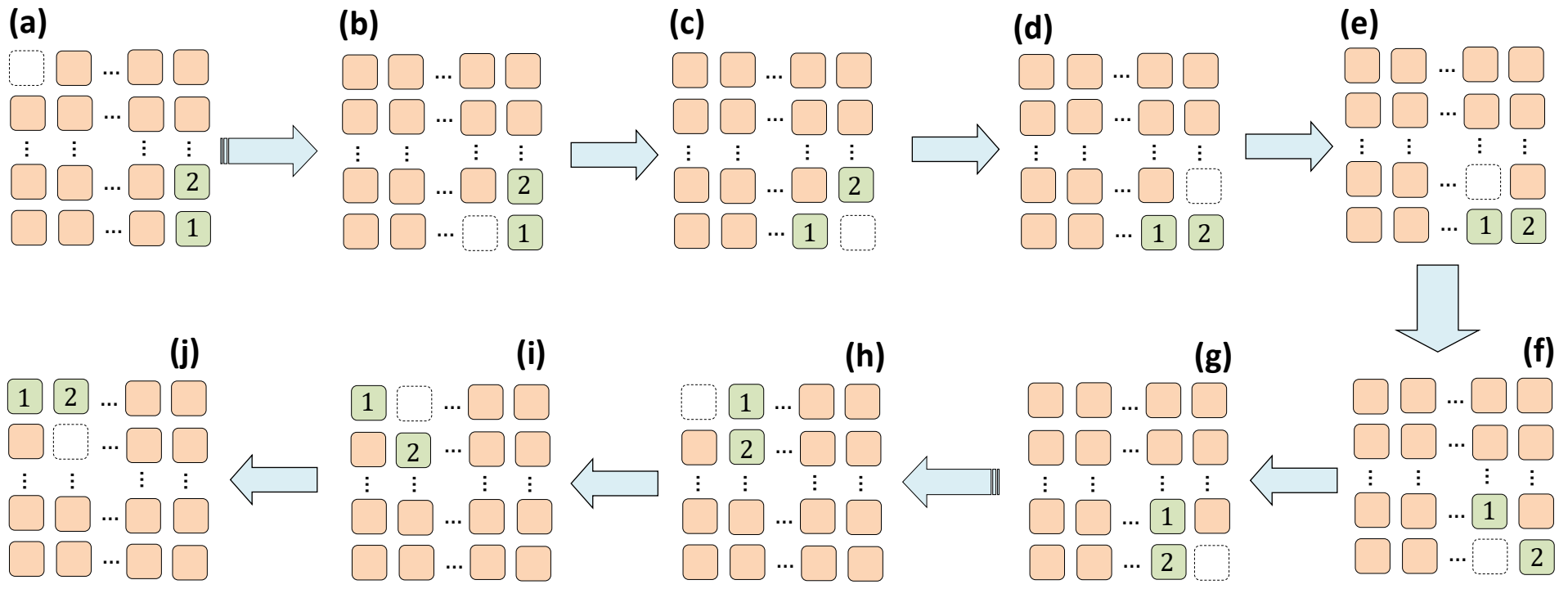

**Figure 3.** *Placing grouped pebbles using a snake.* The situation from Figure 2 is solved by grouping pebbles 1 and 2 into a snake, which is then moved as a whole from its original location in bottom right corner to the goal position in the upper left corner. The process consumes $12n + O(1)$ which is approximately $4n$ better than the original approach that places pebbles individually.

Notice that each pebble itself forms a trivial snake of size 1. Composed movements of a snake *horizontally*, *vertically*, and *diagonally* can be defined analogically as in the case of a single pebble. If fact, they are generalizations of composed movements for single pebble. It is always assumed that the vacant position is in front of the head of snake in the direction of the intended movement. In such a setup, the snake can move forward by one position. The vacant position then needs to be moved around the snake in front of its head again to allow the next movement forward. See Figure 4 for illustration of composed movements for snakes (movements for a snake of length 2 are shown; it is easy to generalize composed movements for snakes of arbitrary length).

The horizontal and vertical composed movements consume $2k + 3$ moves. The number of moves consumed by the diagonal movement depends on the shape of a snake in the middle section – it is not that easy to express. However, if we need to move a snake of length 2 diagonally forward following the shape from Figure 4, then it consumes 10 moves.

Unfortunately it is rarely the case that a group of pebbles in some distant from goal location forms a snake. Even it is not that frequent that pebbles which are to be placed consecutively are close to each other. Hence, to take the advantage of moving a group of pebbles as a snake we need first to form a snake of them. This is however not for free as a number of moves are necessary to form a snake. Thus, it is advisable to consider whether forming a snake is worthwhile. Moreover, there are many ways how to form a snake while each may be of different cost in terms of the number of moves.

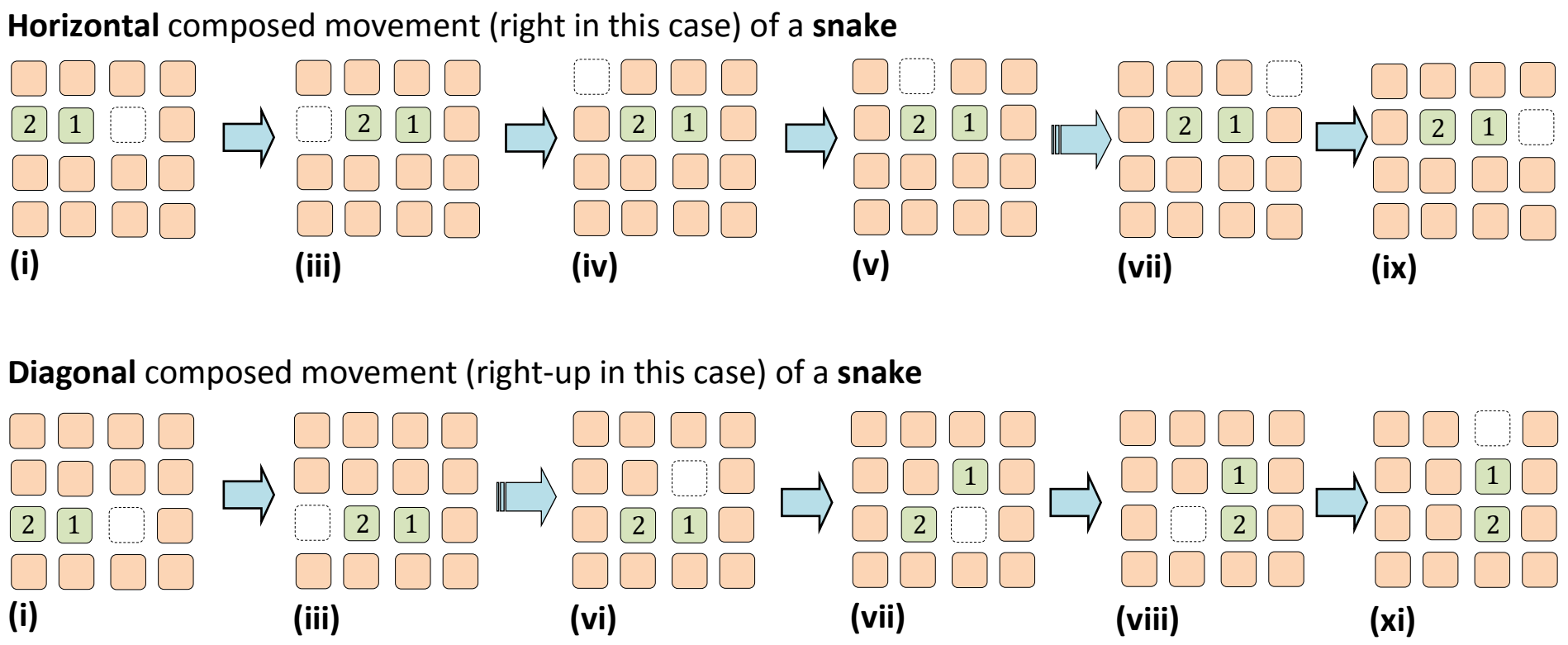

**Figure 4.** *Composed movements of a snake of length 2*. The horizontal and diagonal composed movements of a snake of length 2 are shown. Other cases as well as generalization for snakes of arbitrary length are straightforward.

Generally, the simplest way is to move one pebble to the other or vice versa in order to form a snake of length 2. It is known by using above calculations what number of moves is consumed by moving a snake as well as what number of moves are consumed by moving a pebble towards other pebble. Hence, it is easy to estimate the cost of using a snake in either of both ways as well as the cost of not using it at all in terms of the number of moves. Thus, it is possible to choose the most efficient option. This is another core idea of our new algorithm.

## 4.2. *A 'Snake' Based Algorithm*

Our new algorithm for solving the $(n^2 - 1)$-puzzle will use snakes of length 2. The algorithm proceeds in the same way as the original algorithm of Parberry [13]. That is, pebbles are placed into the first row and then into the first column and after the first row and the first column are finished the task is reduced to the puzzle of the same type but smaller (namely, the task is reduced to solve the $((n-1)^2 - 1)$-puzzle). The trivial case of the 8-puzzle on a grid of the size 3×3 is again solved by the A* algorithm [18].

Along the solving process, the concept of snakes is used to move pebbles in a more efficient way. The basic idea is to make an estimation whether it will be beneficial to form a snake of two pebbles that are about to be placed. If so then a snake is formed in one of the two ways – the first pebble is moved towards the second one or vice versa – the better option according to the estimations is always chosen. If forming a snake turns out not to be beneficial then pebbles are moved in the same way as in the case of the original algorithm; that is, one by one.

**Algorithm 2.** *The main function of a new algorithm for solving the* $(n^2-1)$*-puzzle.* The function for producing a sequence of moves for placing two consecutive pebbles using snakes (if using snakes turns out to be beneficial) is shown.

**function** *Place-Pebbles*$(x_n, y_n, x_n^+, y_n^+, p, q)$: **pair**
/* A function that produces a sequence of moves for placing two consecutive pebbles
with respect to the order of placement. The new configuration is returned in a return value.
Parameters: $x_n, x_y$ - a current configuration of pebbles in the grid,
$x_n^+, y_n^+$ - a goal configuration of pebbles in the grid,
$p, q$ - two consecutive pebbles that will be placed. */

$c \leftarrow \text{cost}_1(x_n, y_n, x_n^+, y_n^+)(p,q)$
$e_{p,q} \leftarrow \text{estimate}_{\text{snake}}(x_n, y_n, x_n^+, y_n^+)(p,q)$
$e_{q,p} \leftarrow \text{estimate}_{\text{snake}}(x_n, y_n, x_n^+, y_n^+)(q,p)$
**if** $\min\{e_{p,q}, e_{q,p}\} < 1.2c$ **then**
  **if** $e_{p,q} < e_{q,p}$ **then**
    **let** $(i,j)$ be a position such that $|i - x_n(p)| + |j - y_n(p)| = 1$
    $(x_n, y_n) \leftarrow$ *Move-Vacant*$(x_n, y_n, i, j)$
    $d_{min} \leftarrow \min\{|i' - x_n(p)| + |j' - y_n(p)| \,|\, i', j' \in \mathbb{N} \wedge |i' - x_n(q)| + |j' - y_n(q)| = 1\}$
    **let** $(i,j)$ be a position such that $|i - x_n(p)| + |j - y_n(p)| = d_{min}$ adjacent to $q$
    $(x_n, y_n) \leftarrow$ *Place-Pebble*$(x_n, y_n, i, j, p)$
  **else**
    **let** $(i,j)$ be a position such that $|i - x_n(q)| + |j - y_n(q)| = 1$
    $(x_n, y_n) \leftarrow$ *Move-Vacant*$(x_n, y_n, i, j)$
    $d_{min} \leftarrow \min\{|i' - x_n(q)| + |j' - y_n(q)| \,|\, i', j' \in \mathbb{N} \wedge |i' - x_n(p)| + |j' - y_n(p)| = 1\}$
    **let** $(i,j)$ be a position such that $|i - x_n(q)| + |j - y_n(q)| = d_{min}$ adjacent to $p$
    $(x_n, y_n) \leftarrow$ *Place-Pebble*$(x_n, y_n, i, j, q)$
  **let** $s = [p, q]$ be a snake {actually $p$ and $q$ form a snake at this point}
  **let** $\pi$ be a shortest path from $(x_n(p), y_n(p))$ to $(x_n^+(p), y_n^+(p))$ such that
    $\pi[|\pi| - 1] = (x_n^+(q), y_n^+(q))$ and $\pi$ does not intersect any position
    containing already placed pebble
  **for** $k = 1, 2, \ldots, |\pi| - 1$ **do**
    $(x_n, y_n) \leftarrow$ *Snake-Composed-Movement*$(x_n, y_n, \pi[k], \pi[k+1], s)$
    {when vacant position is moved it should avoid already placed pebbles}
**else**
  $(x_n, y_n) \leftarrow$ *Place-Pebble*$(x_n, y_n, x_n^+, y_n^+, p)$
  $(x_n, y_n) \leftarrow$ *Place-Pebble*$(x_n, y_n, x_n^+, y_n^+, q)$
**return** $(x_n, y_n)$

Let $\text{estimate}_{\text{snake}}(x_n, y_n, x_n^+, y_n^+): \Omega_n \times \Omega_n \longrightarrow \mathbb{N}_0$ is a functional that estimates the number of moves necessary to place a given two pebbles using the snake like motion. More precisely, $\text{estimate}_{\text{snake}}(x_n, y_n, x_n^+, y_n^+)(p,q)$ is the estimation of the number of moves necessary to form a snake by moving pebble $p$ towards $q$ and to place the formed snake into the goal location where $x_n, y_n$ and $x_n^+, y_n^+$ denote the current and the goal configurations respectively. It is an admissible heuristic if considered in term of A*. Notice, that $\text{estimate}_{\text{snake}}(x_n, y_n, x_n^+, y_n^+)$ can be calculated as sum of distances between several sections multiplied by number of moves needed to travel a unit of distance in that section. However, as different shapes of snake may occur, this calculation may not be exact. Next, let $\text{cost}_1(x_n, y_n, x_n^+, y_n^+): \Omega_n \times \Omega_n \longrightarrow \mathbb{N}_0$ be a functional that calculates exact number of moves necessary to place given two pebbles individually. As the case of individual pebbles is not distorted by any irregularities (such as different shapes as in the case of snake) the number of moves can be calculated exactly – again it is the sum of distances between given sections multiplied by the number of moves needed to travel unit distance in the individual sections.

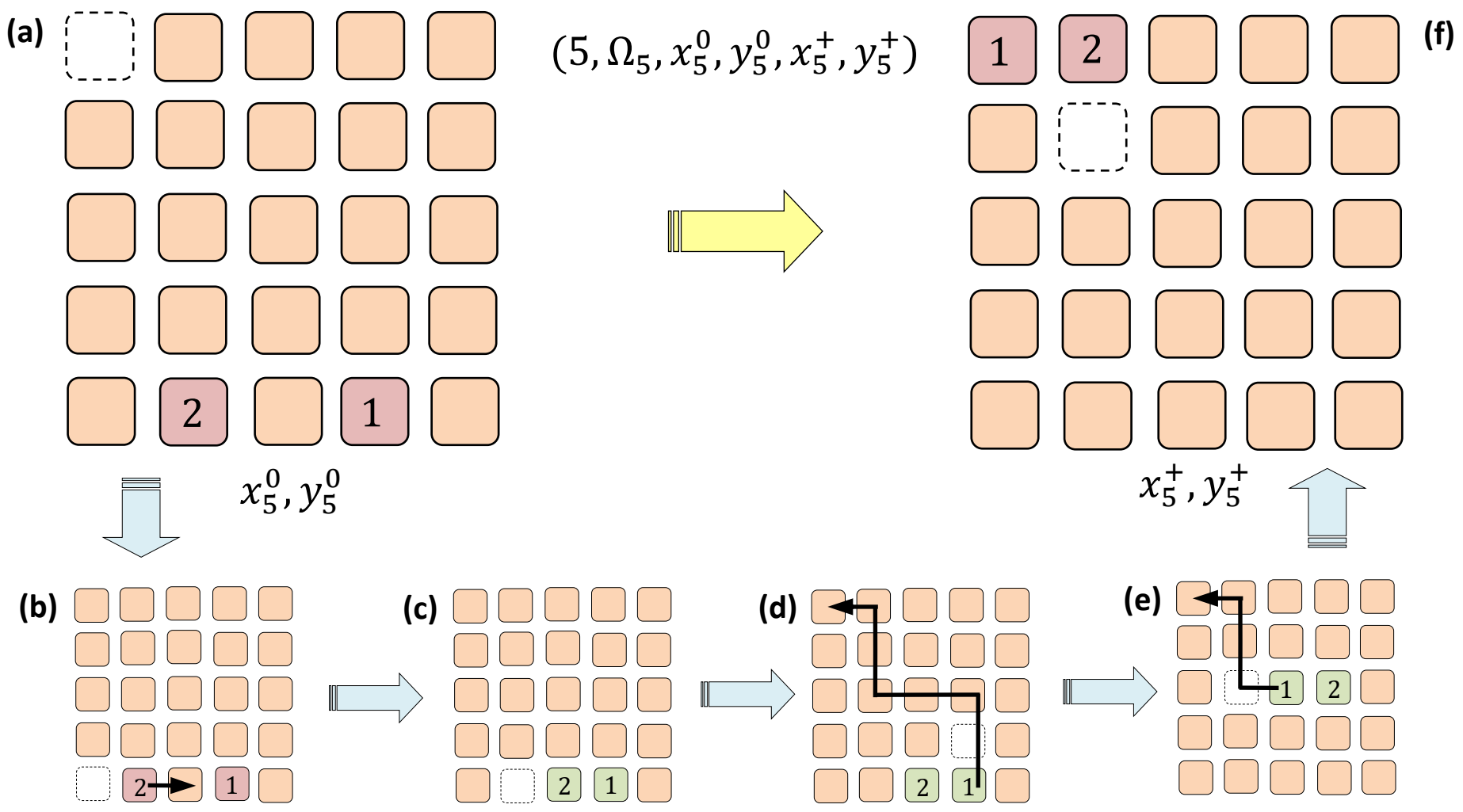

**Figure 5.** *Illustration of snake formation.* A snake will be formed by moving pebble 2 towards pebble 1 and then the whole snake will move to its goal location. The other way of forming a snake is to move pebble 1 towards pebble 2 and then to move the whole snake.

A preliminary experimental evaluation has shown that it suitable to use the following decision rule: if $\min\{\text{estimate}_{\text{snake}}(x_n, y_n, x_n^+, y_n^+)(p, q), \text{estimate}_{\text{snake}}(x_n, y_n, x_n^+, y_n^+)(p, q)\} < 1.2\text{cost}_1(x_n, y_n, x_n^+, y_n^+)(p, q)$ holds then it is tried to form a snake in the better of two ways and to compare the number of moves when snake is used with $\text{cost}_1(x_n, y_n, x_n^+, y_n^+)(p, q)$. If snake is still better then it is actually used to produce sequence of moves into the solution. Otherwise, the original way of placement of pebbles one by one is used.

The main function *Place-Pebbles* for placing a pair of pebbles using snake like motions is shown using pseudo-code as Algorithm 2. It is supposed that the function is used within the main loop of the solving algorithm (Algorithm 1). Several primitives, which all gets current configuration of pebbles as its first two parameters, are used within Algorithm 2: a function *Move-Vacant* moves the vacant position to a specified new location; a function *Place-Pebble* implements the pebble placement process from the original algorithm of Parberry – here it is used as generic procedure to move pebble from one position to another. Finally, a *Snake-Composed-Movement* is a function that implements composed movements of a specified snake; two positions are specified – the current position of the head of snake and the new position for the head. It is also assumed that movement of the snake does not interfere with already placed pebbles. An example of snake formation and its placement is shown in Figure 5.

### 4.3. ***Discussion on Longer Snakes***

We have also considered usage of snakes of length greater than 2. However, certain difficulties preclude using them effectively. There are many more options how to form a snake of length greater than 2. In the case of length $k$, there are at least $k!$ basic options how a snake can be formed (the order of pebbles is determined and then the snake collects pebbles in this order). Moreover, those do not include all the options (for example, it may be beneficial to form two snakes instead of a long one and so on). Therefore considering all the options and choosing the best one is computationally infeasible. Hence, using snakes of length 2 seems to be a good trade-off.

### 4.4. ***Theoretical Analysis***

Although our new algorithm produces locally better sequence of moves for placing a pair of pebbles, it may not be necessarily better globally. Consider that different way of placing the pair of pebbles rearranges other pebbles differently as well, which may influence subsequent movements. Hence, theoretical analysis is quite difficult here. To evaluate the benefit of the new technique in a more realistic manner, we need some experimental evaluation. Nevertheless, theoretical analysis of worst cases can be done at least to get basic insight.

It has been shown that the original algorithm can always find a solution of the length at most $5n^3 + \frac{9}{2}n^2 + \frac{19}{2}n - 89$; that is, $5n^3 + \mathcal{O}(n^2)$ (precisely $5n^3 + 4.5n^2$) [13].

***Proposition 1 (Worst-case Solution Length).*** Our improved algorithm based on snakes can always produce a solution to a given instance of the $(n^2 - 1)$-puzzle of the length of at most $\frac{14}{3}n^3 + \mathcal{O}(n^2)$ (precisely $\frac{14}{3}n^3 + 14n^2$). ■

**Proof.** The worst situation for the worst case calculation of the number of steps using snake-based algorithm we are about to present occurs when the two pebbles – let us denote them $p$ and $q$ – are located in the last row or column. In such a case, we need $14n + \mathcal{O}(1)$ moves in the worst case. Without loss of generality let us suppose both pebbles $p$ and $q$ to be placed in the last row while $p$ is in the first column and $q$ is in the last column. Exactly it is needed: at most $2n - 1$ moves to move the vacant position near $q$; then at most $5(n -$

1) moves to move $q$ towards $p$ which forms a snake; and finally $7n + \mathcal{O}(1)$ moves to relocate the snake into the first row of the grid. These calculations relies on the fact that relocation of the vacant position can be executed by sliding pebbles along a path connecting the source with the goal of the vacant position where each step requires single move. Moving single pebble requires moving vacant position around the pebble to free the position in front of the pebble so moving the pebble one step forward requires 5 moves. Finally, moving the snake is executed in a similar way but the vacant position needs to be moved around the snake, which requires 7 moves.

The algorithm needs to place $n - 1$ pairs of pebbles and one pebble individually in the row. Observe that moving one pebble individually to its goal position requires at most $8n$ moves if snake like movement is used.

Hence, the first row and the first column requires at most $14n^2 + c_1 n + c_0$ moves where $c_0, c_1 \in \mathbb{R}$ with $c_0, c_1 \geq 0$. Let $M(n)$ denotes number of moves needed to solve the $(n^2 - 1)$-puzzle of size $n \times n$. After placing pebbles in the first row and the first columns we have the puzzle of the same type but smaller (the $(n-1)^2 - 1$ puzzle is obtained by omitting the finished row and column) which can used for bounding the number of moves by the following recurrent inequality: $M(n) \leq M(n-1) + 14n^2 + c_1 n + c_0$. The solution of this inequality is $M(n) = \frac{14}{3} n^3 + \mathcal{O}(n^2)$. ■

***Proposition 2 (Worst-case Time Complexity).*** Our new algorithm based on snakes has the worst case time complexity of $\mathcal{O}(n^3)$. ■

**Proof.** The total time consumed by calls of *Move-Vacant* and *Place-Pebble* is linear in the number of moves that are performed. The time necessary to find shortest path avoiding already placed vertices is linear as well since the path has always a special shape that is known in advance (diagonal followed by horizontal or vertical). There is no need to use any path-search algorithm.

Time necessary for calculating $\text{estimate}_{\text{snake}}$ is at most the time necessary to finish the call of *Place-Pebble*, that is, linear in the number of moves again.

Finally, we need to observe that the call of *Snake-Composed-Movement* consumes time linear in the number of moves again since first the shortest path of the special shape needs to be found and then a snake needs to be moved along the path. ■

## 5. Application of 'Snakes' in Cooperative Path-Finding

Promising theoretical and preliminary experimental results from the application of the idea of 'snakes' in solving $(n^2 - 1)$-puzzle inspired us to extend the idea to a closely related problem of *cooperative path-finding* (CPF). The task in cooperative path-finding is to relocate a set of abstract robots over a given undirected graph in a non-colliding way so that each robot eventually reaches its goal vertex [23]. Similarly as in $(n^2 - 1)$-puzzle robots can move into unoccupied vertex while no other robot is allowed to enter the same target vertex at the same time. The natural requirement in CPF is that at least one vertex is empty in the input CPF instance to allow robots to move. Unlike the situation in $(n^2-1)$-puzzle, CPF allows multiple robots to move simultaneously provided there are multiple vacant vertices.

The $(n^2-1)$-puzzle is thus clearly a special variant of CPF where there is only one unoccupied vertex in the graph and the graph, where pebbles (robots) move, has a special structure of the 4-connected grid. A possible application of snakes in CPF is further supported by the fact that several polynomial-time rule-based algorithms that address CPF such as *BIBOX* [26], *PUSH-and-SWAP* [11], and *PUSH-and-ROTATE* [42] relocate robots one by one over the graph towards their goal locations. That is, in the same way as it is done in the algorithm of *Parberry*. Hence, these algorithms are candidates for integrating snake movements into their solving process.

## 5.1. *Cooperative Path Finding Formally*

Cooperative path-finding takes place over an undirected graph $G=(V,E)$ where $V=\{v_1,v_2,\dots,v_n\}$ is a finite set of vertices and $E\subseteq\binom{V}{2}$ is a set of edges. The configuration of robots over the graph is modeled by assigning them vertices of the graph. Let $R=\{r_1,r_2,\dots,r_\mu\}$ be a finite set of *robots*. Then, a configuration of robots in vertices of graph $G$ will be fully described by a *location* function $\alpha: R\longrightarrow V$; the interpretation is that an robot $r\in R$ is located in a vertex $\alpha(r)$. At most one robot can be located in a vertex; that is $\alpha$ is a uniquely invertible function. A generalized inverse of $\alpha$ denoted as $\alpha^{-1}: V\longrightarrow R\cup\{\bot\}$ will provide us a robot located in a given vertex or $\bot$ if the vertex is empty.

**Definition 7** (***Cooperative Path-Finding***). An instance of *cooperative path-finding* problem (CPF) is a quadruple $\Sigma=[G=(V,E),R,\alpha_0,\alpha_+]$ where location functions $\alpha_0$ and $\alpha_+$ define the initial and the goal configurations of a set of robots $R$ in $G$ respectively. □

The dynamicity of the model assumes a discrete time divided into time steps. A configuration $\alpha_i$ at the $i$-th time step can be transformed by a transition action which instantaneously moves robots in the non-colliding way to form a new configuration $\alpha_{i+1}$. The resulting configuration $\alpha_{i+1}$ must satisfy the following *validity conditions*:

(i) $\forall r\in R$ either $\alpha_i(r)=\alpha_{i+1}(r)$ or $\{\alpha_i(r),\alpha_{i+1}(r)\}\in E$ holds (4)
(robots move along edges or not move at all),

(ii) $\forall r\in R\ \ \alpha_i(r)\neq\alpha_{i+1}(r)\Rightarrow\alpha_i^{-1}(\alpha_{i+1}(r))=\bot$ (5)
(robots move to vacant vertices only), and

(iii) $\forall r,s\in A\ \ r\neq s\Rightarrow\alpha_{i+1}(r)\neq\alpha_{i+1}(s)$ (6)
(no two robots enter the same target/unique invertibility of the resulting configuration).

The task in cooperative path finding is to transform $\alpha_0$ using above valid transitions to $\alpha_+$. An illustration of CPF and its solution is depicted in Figure 6.

**Definition 8** ***(Solution).*** A *solution* of a *makespan* $m$ to a cooperative path finding instance $\Sigma = [G, R, \alpha_0, \alpha_+]$ is a sequence of configurations $\vec{s} = [\alpha_0, \alpha_1, \alpha_2, \dots, \alpha_m]$ where $\alpha_m = \alpha_+$ and $\alpha_{i+1}$ is a result of valid transformation of $\alpha_i$ for every $= 1,2, \dots, m-1$. □

The number $|\vec{s}| = m$ is a *makespan* of solution $\vec{s}$. It is known that deciding whether there exists a solution of CPF of a given makespan is $NP$-complete [17, 30].

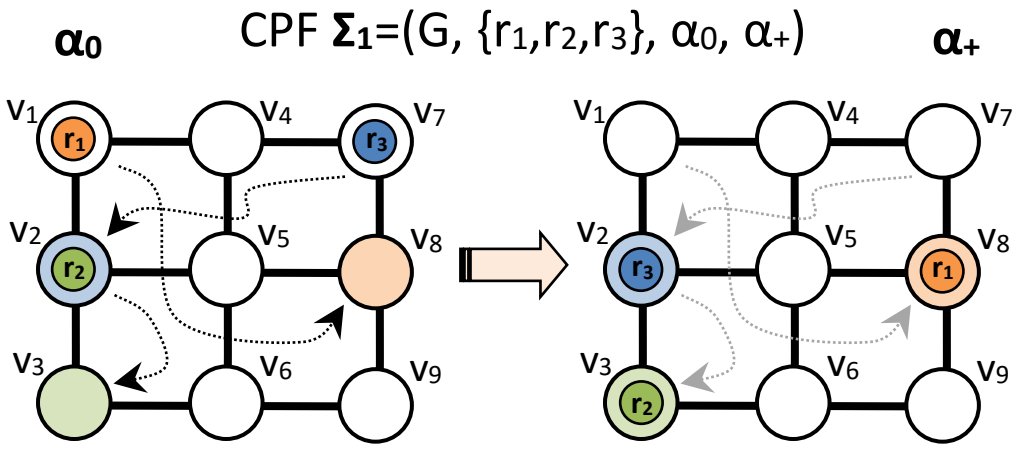

| $\vec{s}$ | $\alpha_0$ | $\alpha_1$ | $\alpha_2$ | $\alpha_3$ | $\alpha_4 = \alpha_+$ |
|---|---|---|---|---|---|
| $r_1$ | $v_1$ | $v_1$ | $v_2$ | $v_5$ | $v_8$ |
| $r_2$ | $v_2$ | $v_3$ | $v_3$ | $v_3$ | $v_3$ |
| $r_3$ | $v_7$ | $v_4$ | $v_4$ | $v_1$ | $v_2$ |

**Figure 6.** *An example of cooperative path-finding problem (CPF).* Three robots $r_1, r_2$, and $r_3$ need to relocate from their initial positions represented by $\alpha_0$ to goal positions represented by $\alpha_+$. A solution of makespan 4 is shown.

## 5.2. *Introducing 'Snakes' into the BIBOX Algorithm*

We will briefly recall basics of the *BIBOX* algorithm before the improvement with snakes will be integrated into it. The comprehensive description and evaluation of the algorithm is given in [33] to which we refer the reader for further details. The algorithm is designed for CPFs over *bi-connected graphs* with at least two unoccupied vertices (modifications for single unoccupied vertex exist as well [29]).

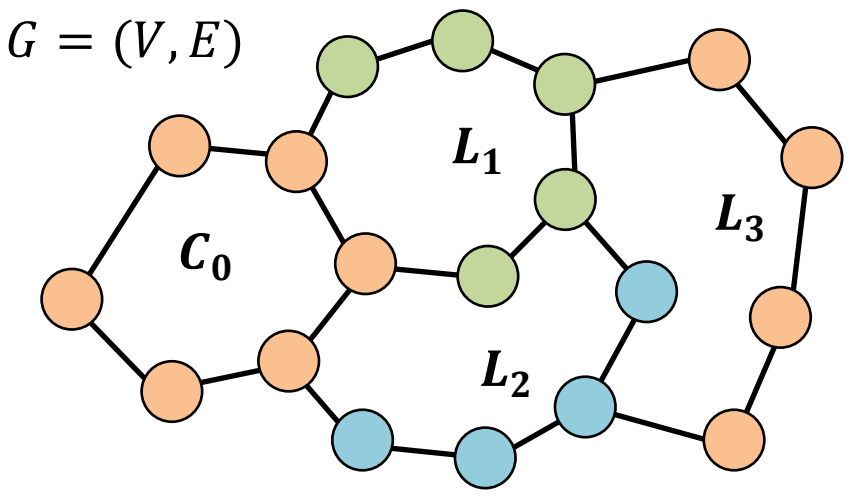

**Figure 7.** *Example of* ***bi-connected graph***. An ear decomposition is illustrated. The graph can be constructed by starting with cycle $C_0$ and by gradually adding ears $L_1$, $L_2$, and $L_3$.

**Definition 9** ***(connected graph).*** An undirected graph $G = (V, E)$ is *connected* if $|V| \geq 2$ and for any two vertices $u, v \in V$ such that $u \neq v$ there is an undirected path connecting $u$ and $v$. □

**Definition 10** **(*bi-connected graph, non-trivial*).** An undirected graph $G = (V, E)$ is *bi-connected* if $|V| \geq 3$ and the graph $G' = (V', E')$, where $V' = V \setminus \{v\}$ and $E' = \{\{u, w\} | u, w \in V \wedge u \neq v \wedge w \neq v\}$, is connected for every $v \in V$. A bi-connected graph not isomorphic to a cycle will be called *non-trivial* bi-connected graph. □

Observe that, if a graph is bi-connected, then every two distinct vertices are connected by at least two *vertex disjoint paths* (equivalently, there is a cycle containing both vertices; only internal vertices of paths are considered when speaking about vertex disjoint paths - vertex disjoint paths can intersect in their *start points* and *endpoints*). An example of bi-connected graph is shown in Figure 7.

An algorithmically important property of bi-connected graphs is that every bi-connected graph can be constructed from a cycle by adding sequence of ears to the currently constructed graph [36, 40, 41]. The *BIBOX* algorithm is substantially based on this property. Consider a graph $G = (V, E)$; the new ear with respect to $G$ is a sequence $L = [u, w_1, w_2, \dots, w_h, v]$ where $h \in \mathbb{N}_0$, $u, v \in V$ such that $u \neq v$ (called *connection vertices*) and $w_i \notin V$ for $i = 1,2, \dots, h$ ($w_i$ are fresh vertices). The result of the addition of the ear $L$ to the graph $G$ is a new graph $G' = (V', E')$ where $V' = V \cup \{w_1, w_2, \dots, w_h\}$ and either $E' = E \cup \{\{u, v\}\}$ in the case of $h = 0$ or $E' = E \cup \{\{u, w_1\}, \{w_1, w_2\}, \dots, \{w_{h-1}, w_h\}, \{w_h, v\}\}$ in the case of $h > 0$. Let the sequence of ears together with the initial cycle be called an *ear decomposition* of the given bi-connected graph. Again, see Figure 7 for illustrative example.

**Lemma 1** ***(ear decomposition)*** [36, 40, 41]**.** Any bi-connected $G = (V, E)$ graph can be obtained from a cycle by a sequence of operations of adding an ear. ■

The important property of the construction of a bi-connected graph according to its ear decomposition is that the currently constructed graph is bi-connected at every stage of the construction. The algorithm for solving CPFs over bi-connected graphs can proceed inductively according to the ear decomposition by arranging robots into individual ears – after finishing placement of robots into an ear, the problem reduces to a problem of the same type but on a smaller graph without the currently solved ear.

As the *BIBOX* algorithm has been already thoroughly published, its enhancement with snakes described using pseudo-code has been deferred to Appendix B (Algorithm 3). The idea behind using snakes within the *BIBOX* algorithm is similar as in the case of the algorithm of *Parberry*.

Again, snakes of length 2 are used within the modified *BIBOX* algorithm. Consider that robots $r$ and $s$ are two consecutive robots within the processed ear $L_i$. In the original algorithm, they are moved one by one towards the ear connection vertex and stacked inside the ear by its rotation afterwards; that is, relocation and stacking inside the ear of $r$ and $s$ is done separately. When snake reasoning is used, it is first checked if $r$ and $s$ are close enough to each other before $r$ is relocated towards the ear connection vertex. If it is the case, then $r$ and $s$ are relocated together jointly until $r$ reaches the connection vertex. After such relocation, robot $s$ is next to the connection vertex and can be then stacked into the ear quickly. If robots $r$ and $s$ are too far from each other, then the original relocation of both robots separately is used. The process of relocation of two consecutive robots is implemented by procedure *Move-Robot-Snake* within the pseudo-code of Algorithm 3.

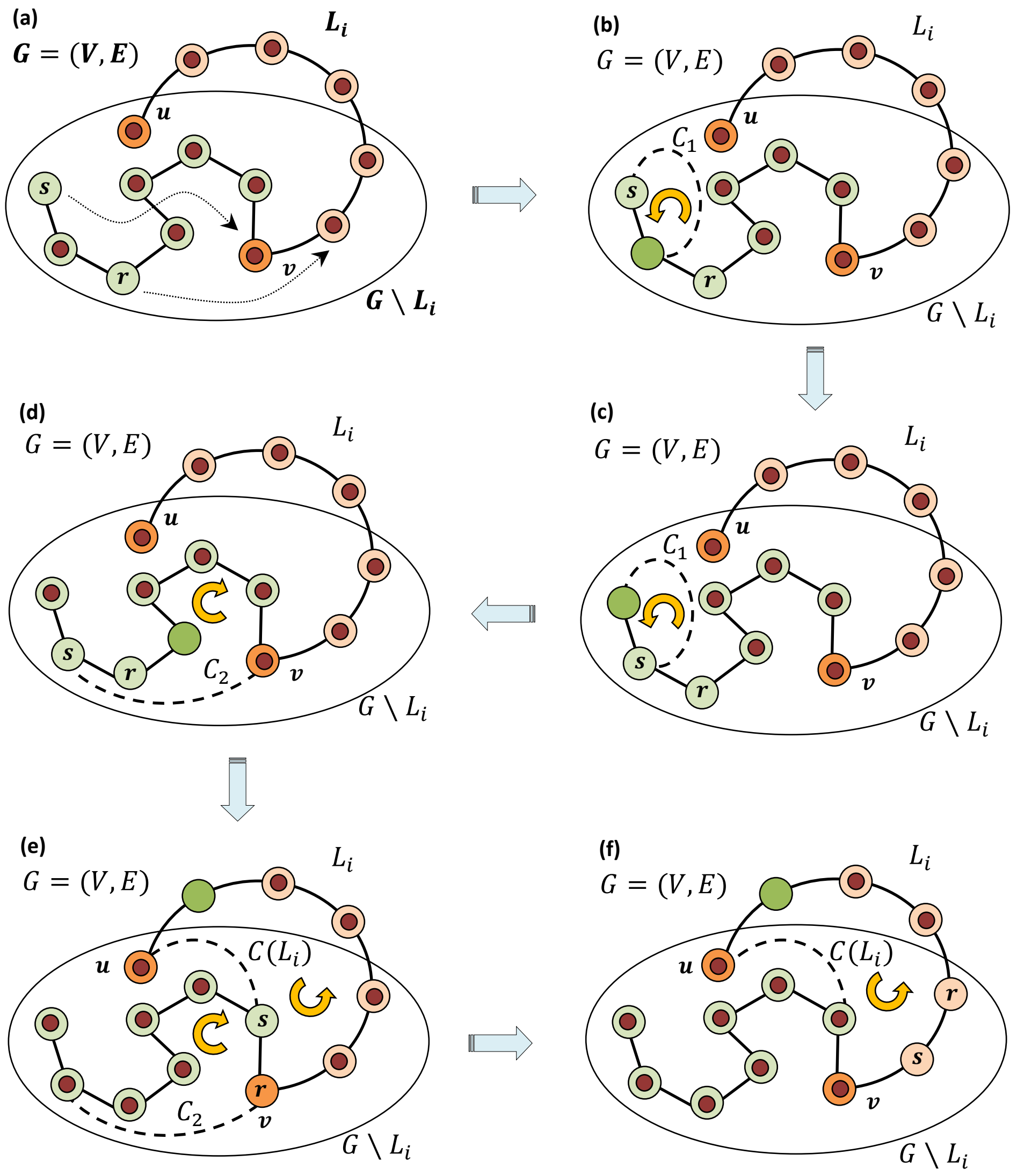

**Figure 8.** *Illustration of using* ***snakes of size 2*** *within the* ***BIBOX*** *algorithm.* A pair of robots $r$ and $s$ needs to be stacked into ear $L_i$ next to each other. They first need to be moved towards ear connection vertex $v$. Shortest paths connecting robot locations with vertex $v$ are depicted. According to distance heuristic it is decided that $r$ and $s$ should be relocated together. Thus, robot $s$ is moved next to $r$ by rotating cycle $C_1$ (stages (b), (c)). Then, $r$ and $s$ moves like a snake jointly by rotating cycle $r$ and $C_2$ until $r$ appears in the ear connection vertex $v$ (stage (d)). Finally, robot $r$ is stacked into ear $L_i$ by rotation of cycle $C(L_i)$ associated with the ear (stage (e)). As robot $s$ has been next to $r$ all the time, it consequently moved to its target vertex during the last rotation as well (stage (f)). Other robots whose identity is irrelevant now are depicted as circles in vertices.

Let us now clarify what does it mean that robots are close enough to each other and what the joint relocation means. When considering if snake based relocation pays-off, a simple distance heuristic is used. The cost of relocation is estimated by the length of shortest path between the original and target location. Let $v$ be an ear connection vertex, $\alpha$ current configuration of robots and let $\text{dist}_G(u, v)$ for $u, v \in G$ denote the shortest path between $u$ and $v$ in $G = (V, E)$. Then robots $r$ and $s$ are relocated jointly if:

$$\text{dist}_G(v, \alpha(r)) + \text{dist}_G(\alpha(r), \alpha(s)) < \text{dist}_G(v, \alpha(r)) + \text{dist}_G(v, \alpha(s)) \quad (7)$$

That is, if relocation of $s$ towards $r$ and relocation of $r$ and $s$ jointly towards $v$ seems to be less costly than relocation of $r$ and $s$ towards $v$ separately.

The original relocation of a robot $r$ within the *BIBOX* algorithm is done by finding a cycle which includes the target vertex and robot $r$. One vertex within the cycle is made unoccupied which enable rotation of the cycle. Robot $r$ is moved towards its target by rotating the cycle until $r$ appears in the target vertex. The original relocation is implemented by *Move-Robot* procedure in the pseudo-code.

The joint relocation of a pair of robots uses the very same idea. First, robot $s$ is moved next to $r$ by the original way of relocating robots (*Move-Robot*). Then a cycle containing the edge, whose endpoints are occupied by $r$ and $s$ respectively, is rotated. The cycle rotated until $r$ reached its target. Throughout the series of rotations of the cycle, robots $r$ and $s$ are preserved to stay next to each other, which eventually means that $s$ is close to its target at the end of joint relocation. The joint relocation is implemented by *Move-Robot-Snake* within the pseudo-code. The illustration of the process of joint relocation of a pair of robots is shown in Figure 8.

## 5.3. *Introducing 'Snakes' into the PUSH-and-SWAP Algorithm*

The *PUSH-and-SWAP* algorithm [11] and its later corrected version *PUSH-and-ROTATE* [42, 43] are more general than *BIBOX* as they are complete for CPFs over arbitrary graphs with at least two unoccupied vertices. We will concentrate here on the original version of *PUSH-and-SWAP* though it is not complete since certain cases are not treated by the algorithm [43] – these cases causing incompleteness however do not affect our study.

The *PUSH-and-SWAP* algorithm proceeds in solving of CPF in a similar way as *BIBOX* and the *Parberry's* algorithm since it also places robots to their goal positions one by one while already placed robots are restricted to move. This characteristic makes the algorithm suitable for improvement via snakes. Moving multiple robots opportunistically together towards their goals may save moves over the process of moving them towards goals separately. This is a hypothesis that we would like to investigate.

We will focus here on two basic operations, on that the algorithm relies, known as *Push* and *Swap*. A robot is always moved towards its goal along a path that connects the current position of the robot with its goal. After the robot reaches its goal it is locked there so the subsequent robot rearrangements can move the robot out of its goal vertex temporarily only (the robot must eventually return to its goal).

The *Push* operation tries to push the robot by one step (by one vertex) towards its goal along the path. If the next vertex on the path is occupied by a robot and is unlocked then *Push* tries to make the next vertex unoccupied by moving the robot out of it.

If the *Push* operation is unsuccessful then it is the turn of the *Swap* operation. The *Swap* operation exchanges positions of the robot being relocated and the next robot on the path. Assume that robot $r$ is relocated towards the goal and the next vertex on the path is occupied by robot $s$. The *Swap* operation will try to move robots $r$ and $s$ to a vertex of the degree at least 3 where $r$ and $s$ are exchanged.

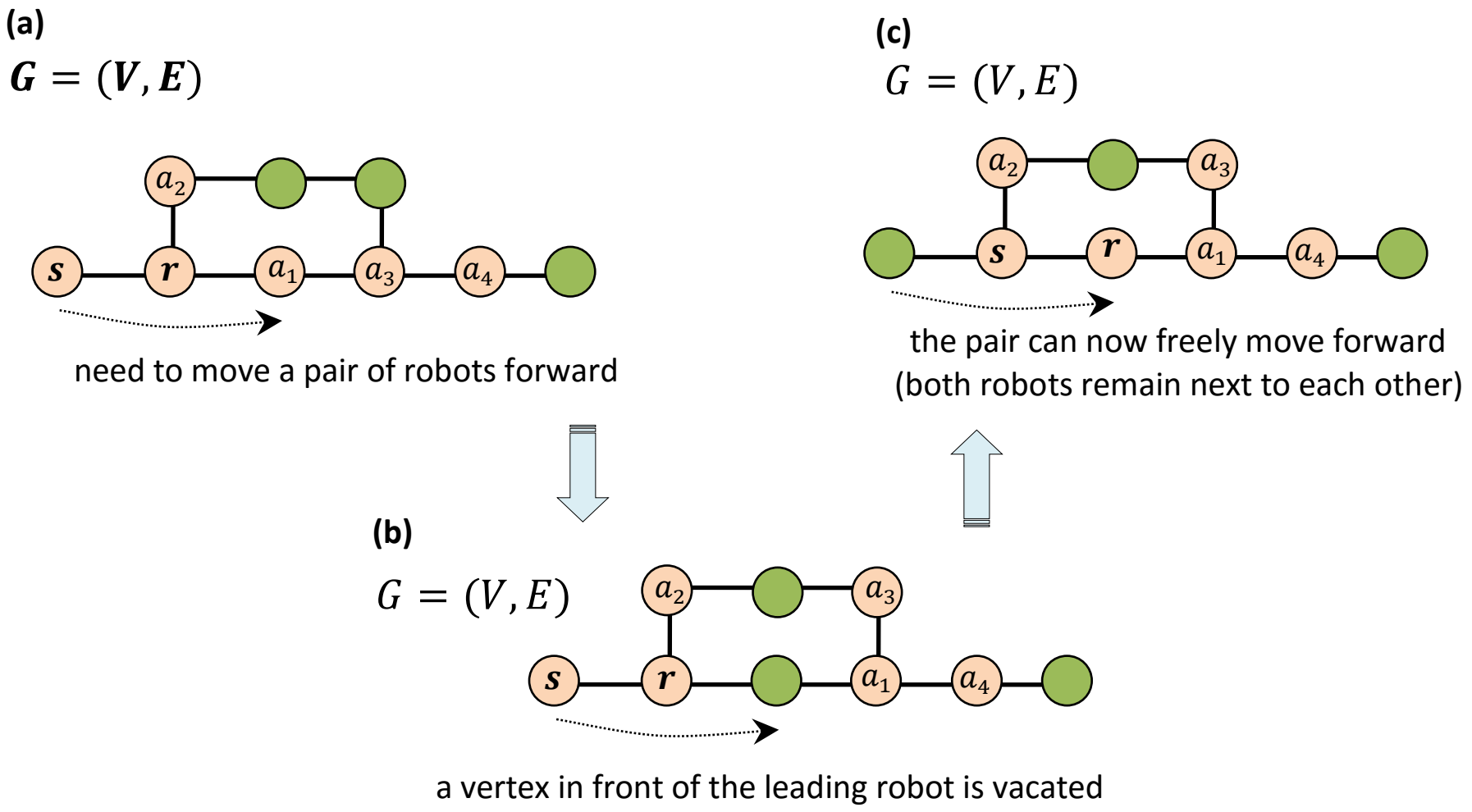

**Figure 9.** *Illustration of the* ***joint relocation*** *of a pair of robots (within the* ***Twin-Push*** *operation).* A pair of robots is moved forward – robot $r$ is the leader followed by robot $s$. After the step forward both robots are next to each other again. The illustrated sequence of moves represent the *Twin-Push* operation.

Assume we have vertex $v$ with $\deg_G(v) \geq 3$. Robots $r$ and $s$ are moved together towards $v$ while $r$ is the leading robot followed by $s$. Vertex locking is disregarded now so robots can be moved out of their locked goals. With $r$ in $v$ and $s$ next to it, the two robots are locked in their positions so the subsequent operations cannot move them out. Then two neighbors of $v$ are vacated. As $r$ and $s$ locked in their positions, we have $r$ with two blanks and $s$ as its neighbors. In such a configuration it is possible to exchange $r$ and $s$.

Robots moved out of their goals during the process preceding the exchange need to be put back. This is done by recording all the moves that precede the exchange at $v$. By executing the recorded moves reversely in the reverse order we restore placement of other robots except $r$ and $s$ which finish swapped at their original positions.

The idea behind introducing snakes into the *PUSH-and-SWAP* algorithm is that moving robots by the *Swap* operation may consume fewer moves if robots are relocated as a pair (snake of length 2) instead of being moved separately (see Figure 9). This is formally justified by following two lemmas and the following calculation (see Figure 10 for illustration of lemmas).

**Lemma 2** ***(exchange of robot).*** A pair of robots can be exchanged at a vertex with at least 2 vacant neighbors by using 6 moves. A triple of robots can be exchanged at a vertex with at least 3 vacant neighbors by using 12 moves. ■

The proof of the first statement is given in [43]; the illustration of the second statement is shown in Figure 11.

**Lemma 3** ***(robot relocation).*** Relocation of a robot along a path of length $k$ while disregarding the locked vertices requires at most $k$ removals of a robot from the next vertex on the path. Relocation of a pair of robots along a path of length $k$ requires at most $k$ removals of a robot from the next vertex on the path as well. ■

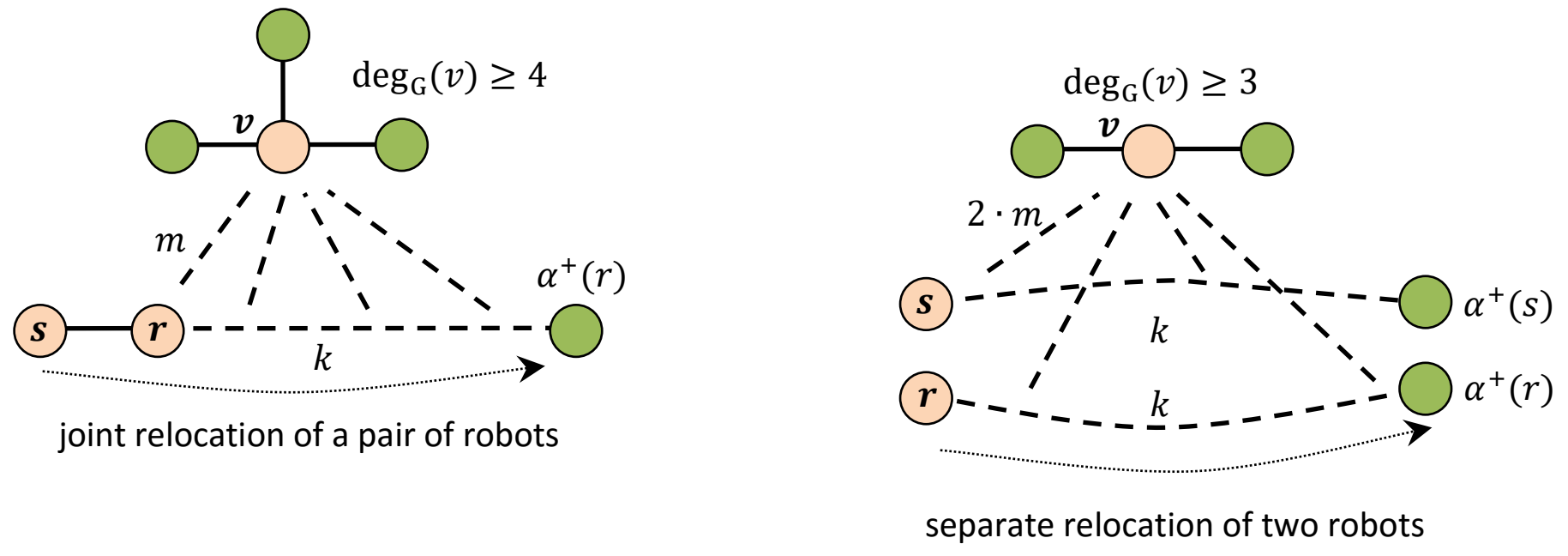

**Figure 10.** *Schematic illustration of the* ***advantage of the joint relocation*** *of the pair of agents over their separate relocation.* Two robots need to traverse path of length $k$ towards the goal of the first robot. We assume that path of length $k$ requires traversing an additional distance $m$ to reach a vertex with enough neighbors where exchange of robots is performed. In case of separate relocation of the pair of robots, moving towards the vertex with enough neighbors may require traversing double distance than in case of joint relocation.

Assume that relocation of two robots needs to traverse paths of length $k$ using the *Swap* operation while reaching vertices of the degree at least three means to travel distance $m$ in total. This means that a vertex is vacated $2 \cdot m + 2 \cdot k$ times and a pair of robots is exchanged $2 \cdot k$ times which is equivalent to $2 \cdot k \cdot 6 = 12 \cdot k$ moves altogether in the worst case.

If the pair of robots is relocated in the similar configuration (the length of path is $k$ and reaching the vertices of the degree at least 4 requires traversing the distance $m$ in total) jointly, then we need to vacate a vertex $m + 3 \cdot k$ times and a triple of robots is exchanged $k$ times which is equivalent to $12 \cdot k$ moves in the worst case. Hence, the difference consists in how many times a vertex is vacated during the process. If $m + 3 \cdot k < 2 \cdot m + 2 \cdot k$ that is, if $k < m$ then relocation of robots jointly should produce fewer moves in the worst case as fewer vertex vacations are needed.

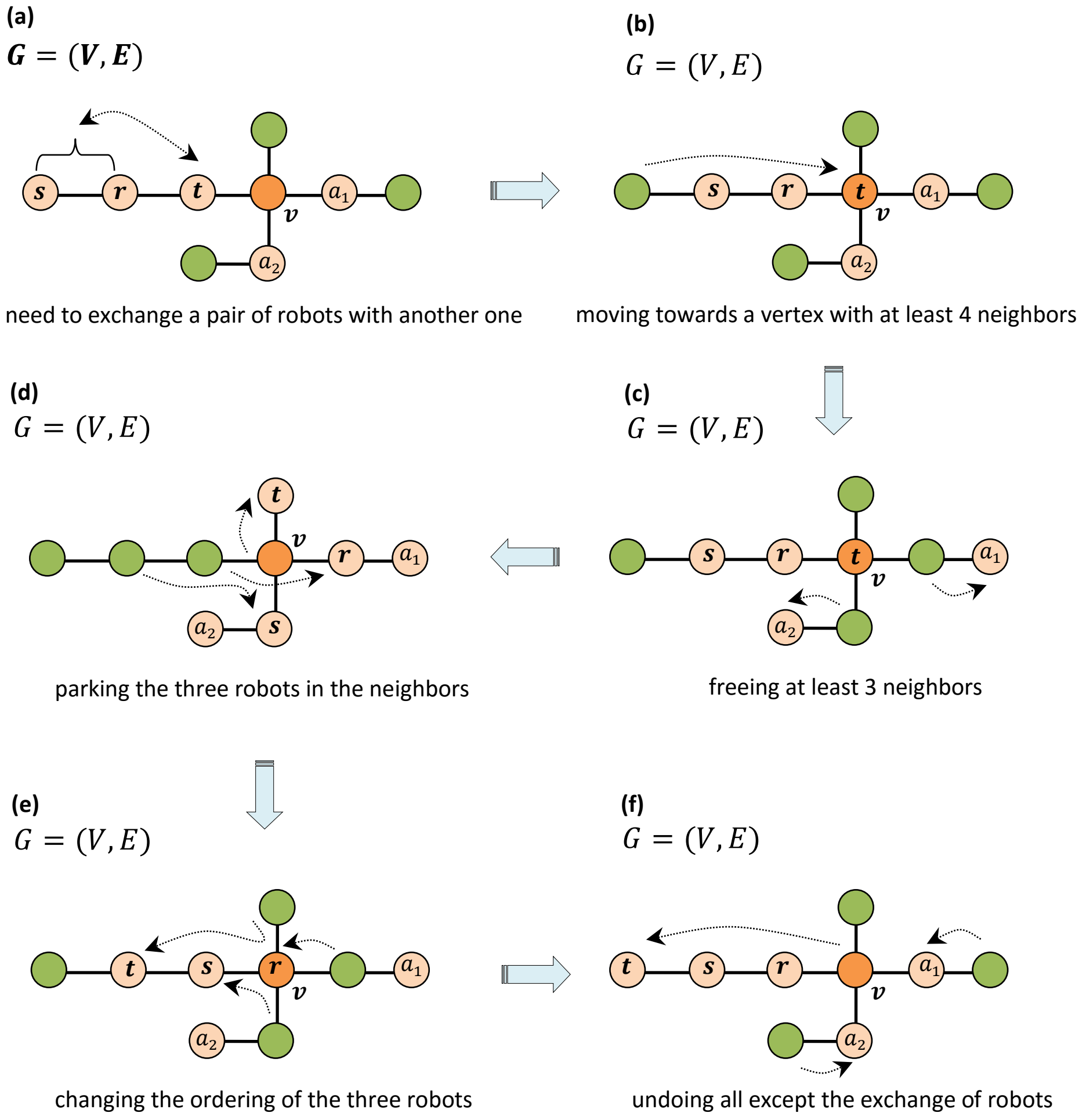

**Figure 11.** *Illustration of the* ***Twin-Swap operation*** *within the snake enhanced PUSH-and-SWAP algorithm. Twin-Swap* is an analogy to the original *Swap* operation for the snake of length two. The original *Swap* operation swaps a pair of robots using a vertex with at least three neighbors while the *Twin-Swap* operation swaps a pair of robots $r$ and $s$ with the third robot $t$ using a vertex with at least four neighbors $v$.

Although the above calculation is based on strong assumptions that may not fully hold in practice it lead us to the design of a version of the *PUSH-and-SWAP* algorithm where robots are relocated together in pairs towards their goals. We suggested an analogy to the *Push* and *Swap* operations called *Twin-Push* and *Twin-Swap* (see Figure 11 for illustration) that instead of moving a single robot, move a pair of robots jointly. The formal description of the *PUSH-and-SWAP* algorithm enhanced with snakes is given using pseudo-code in Appendix C. Let us now describe only the top-level change to the original *PUSH-and-SWAP* algorithm. The detailed implementation of basic operations is described in the appendix.

Instead of placing single robot to its goal, two robots consecutive in some ordering are taken and relocated towards the goal of the former one jointly. The hypothesis is that if a right ordering of robots is taken, then the later robot appears close to its goal after relocation of the pair and only few moves are needed to reach the goal eventually by the latter robot. The (sequential) ordering of robots according to which robots are taken for placement to their goals should satisfy that robots close to each other in the sequence should have goals close to each other in the underlying graph as well. Such a sequential ordering of robots can be obtained by the breadth first search of graph.

However, the joint relocation cannot be applied immediately. First, robots those are consecutive with respect to the given ordering need to be placed next to each other, which requires some extra work and decision-making.

There are three options how to perform the relocation of the two robots:

(i) move the former robot next to the latter one and then relocate robots together towards the goal of the former robot

(ii) move the latter robot next to the former one and then relocate robots together towards the goal of the former robot

(iii) do not use the joint relocation at all

A heuristic decision based on shortest paths calculation can be used to choose one of these options (as suggested in inequality (7)). However, our preliminary experimental evaluation indicated that such a heuristic does not provide acceptable results. The estimation by shortest path does not correspond to the actual number of moves performed by the algorithm (it is a lower bound far from the actual number of moves). This is a significant difference from the *BIBOX* algorithm where this heuristic worked well. Hence, we used simulation by which we are able to calculate exact number of moves consumed by each of the three options and chose that one that consumed fewest moves.

In the case of *PUSH-and-SWAP* algorithm snakes longer than 2 were not considered as more vacant vertices would be needed and the algorithm will become too complex. Also the above decision making will become much more as the number of options grows exponentially with the increasing length of snakes.

## 6. Experimental Evaluation

An experimental evaluation is necessary to explore qualities of our new snake-based improvements to the studied rule-based pebble relocation and cooperative path-finding algorithms. The snake-based improvements to *Parberry's* algorithm as well as its variants within rule-based cooperative path-finding algorithms *BIBOX* and *PUSH-and-SWAP* have been implemented in C++ to make the experimental evaluation possible.

The *Parberry's* algorithm and the *PUSH-and-SWAP* algorithm have been fully re-implemented in C++. The snake-based reasoning within the *BIBOX* algorithm has been integrated into its original C++ implementation [26] where only minor changes to the code needed to be made. Using the C++ language and the same coding style for implementations of all the evaluated algorithms allowed us to ensure equal testing conditions.

A series of tests has been conducted to measure the total number of moves performed by each algorithm in various problem setups. The runtime necessary to solve given instances has been measured too.

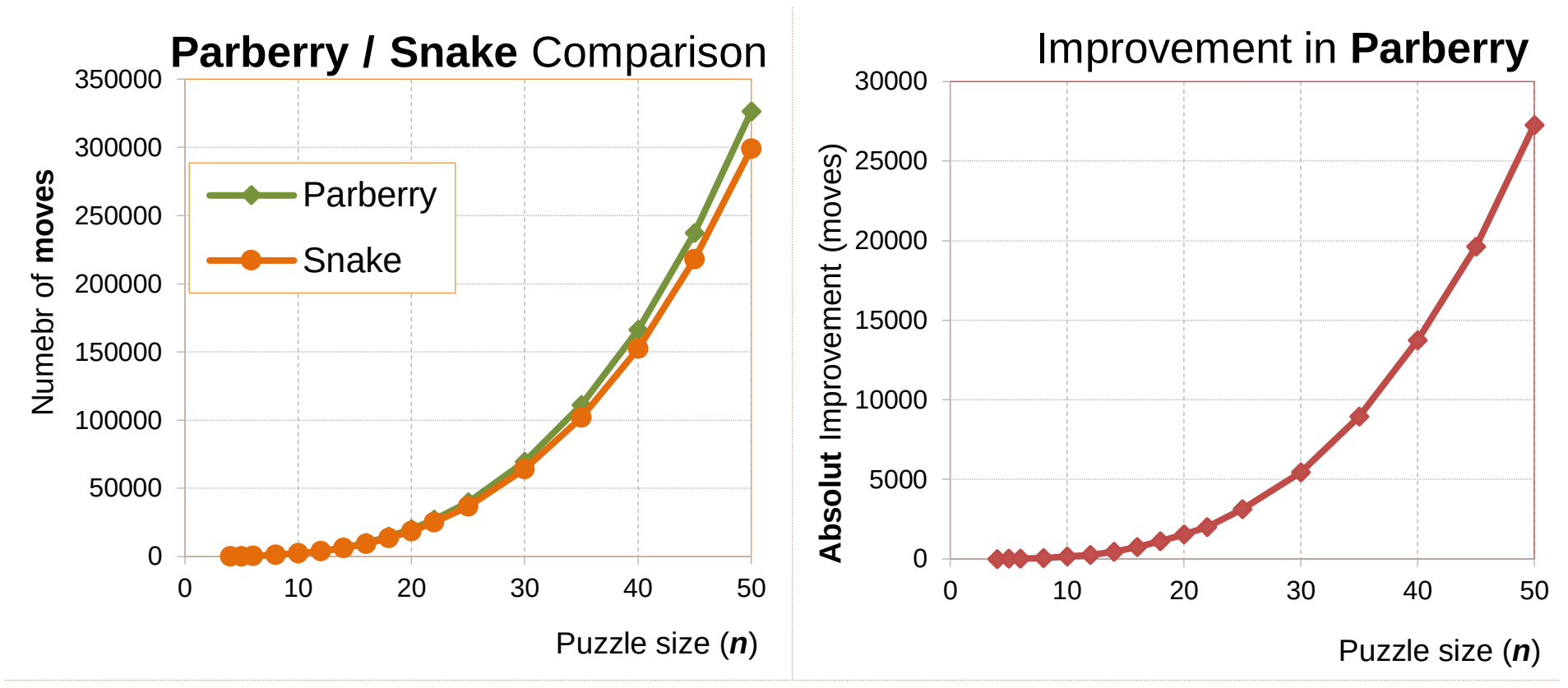

**Figure 12.** *Comparison of the original **Parberry's algorithm** and its **snake-based improvement** in terms of total number of steps.* Comparison has been done for several sizes of the puzzle ranging from 4 to 50. Forty random instances were generated for each size of the puzzle. The average number of moves for both algorithms is shown in the left part. The absolute improvement that can be achieved by using snakes is shown in the right part.

In order to give well readable results we used basic versions of all the evaluated algorithms where sequence of moves with no parallelism is produced as a solution. Further solution improvements that increases parallelism [28, 39, 43] (multiple independent moves can be performed in a single time-step) and removes redundancies [27, 32, 39, 43] were not applied as it is out of scope of this study.

We have three rule-based algorithms and the snake-based improvement of each of them, which makes six algorithms to be tested altogether. Such an extensive experimental setup needs to be well designed to obtain meaningful results.

Hence, we suggested evaluating each snake-improved algorithm against its original variant to see how beneficial the introduction of snakes is. This evaluation has been done on $(n^2 - 1)$-puzzles and in the case of more general CPF algorithms *BIBOX* and *PUSH-and-SWAP* also on instances over randomly generated bi-connected graphs.

The complete C++ source code and raw experimental data are available for download on the website: http://ktiml.mff.cuni.cz/~surynek/research/j-puzzle-2015 to allow full reproducibility of presented results and own experiments with snake-based improvements in tested algorithms.

### 6.1. *Experimental Evaluation of Snakes in the Parberry's Algorithm*

In the case of snake-based improvement to the *Parberry's* algorithm, we have only the upper bound estimation of the total number of steps so far which however does not show that the new algorithm actually produces fewer moves. Thus, a thorough empiric evaluation needs to be done.

| **Relative Improvement** in Parberry's Algorithm | |
|---|---|
| $n$ | **Length Improvement** (%) |
| 4 | 5.58 |
| 5 | 4.79 |
| 6 | 5.57 |
| 8 | 5.75 |
| 10 | 6.34 |
| 12 | 6.51 |
| 14 | 6.59 |
| 16 | 6.66 |
| 18 | 7.29 |
| 20 | 7.51 |
| 22 | 7.54 |
| 25 | 7.49 |
| 30 | 7.84 |
| 35 | 7.84 |
| 40 | 8.07 |
| 45 | 8.26 |
| 50 | 8.26 |

**Table 1.** ***Relative improvement*** *achieved by using snakes in the Parberry's algorithm.* Again, the improvement has been measured for several sizes of the puzzle ranging from 4 to 50. For each size, 40 random instances were generated and the average improvement was calculated.

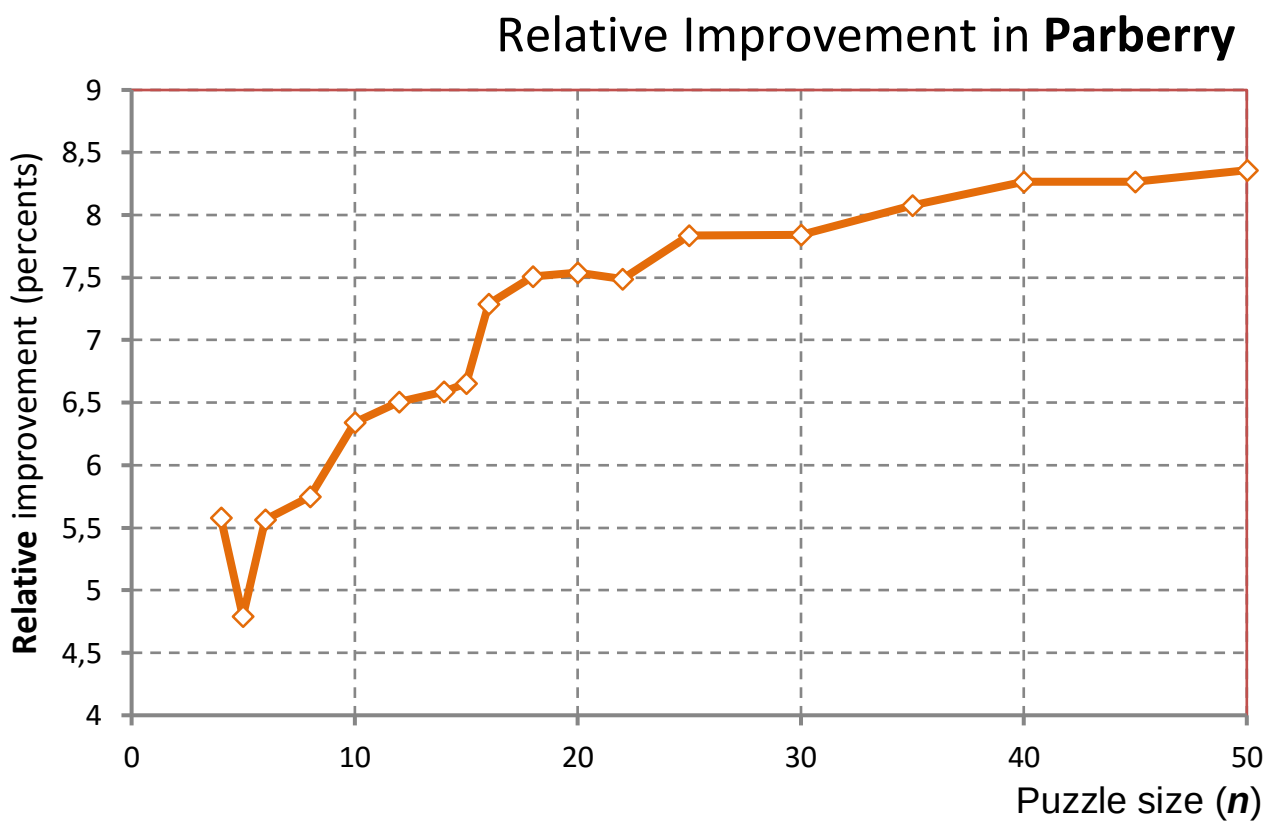

**Figure 13.** *Illustration of the* ***trend*** *in the average improvement.* It can be observed that the relative improvement tends to stabilize between 8% and 9% as instances are getting larger.

Regarding the choice of testing puzzles, we followed the benchmark generation already used in the literature by Korf and Taylor [9] where random instances of the $(5^2 - 1)$-puzzle were used. There is experimental evidence that solving random instances of the $(5^2 - 1)$-puzzle optimally is difficult [9]. Although there is a difference between optimal and sub-optimal solving we consider that these instances are suitable for producing interesting results in a sub-optimal case as well.

A random instance of the $(n^2 - 1)$-puzzle is generated by placing pebbles randomly (with uniform distribution) to remaining unoccupied positions within the initial arrangement. The goal arrangement consists of pebbles ordered linearly in rows of the 4-connected grid forming the environment starting in the first row and the first column and continuing to the last row and last column where unoccupied position is located. Among randomly generated instances those solvable are taken for experiments.

#### 6.1.1. *Competitive Comparison of the Parberry's Algorithm with Snakes on Puzzle Instances*

The competitive comparison of the total number of moves produces by the algorithm of Parberry and its snake-based improvement is shown in Figure 12. The improvement achieved by snake-based approach is illustrated as well. For each size of the instance, average out of 40 random instances is shown.

It is observable that the growth of the number of moves for growing size of the instance is polynomial. Next, it can be observed that snakes achieve a stable improvement, which is proportional to the total number of moves. The more detailed insight into the achieved improvement of the total number of moves is provided in Figure 13 and Table 1. It clearly indicates that the improvement is becoming stable between 8% and 9% with respect to the original algorithm, as instances are getting larger.

#### 6.1.2. *Parberry's Algorithm on Individual Puzzle Instances*

Comparison of the total number of moves on the individual instances of various sizes is shown in Figure 14. These results show that using snakes, even though it is locally a better choice, can lead to global worsening of the solution. This phenomenon sometimes occurs exclusively on small instances. Here it is visible for instances of the size of $4\times4$.

On larger instances, the local benefit of using snakes dominates over any local worsening of the configuration so there is stably significant improvement of 7% and 9%. That is, the improvement is not only observable in average calculated from several instances but also on a single instance.

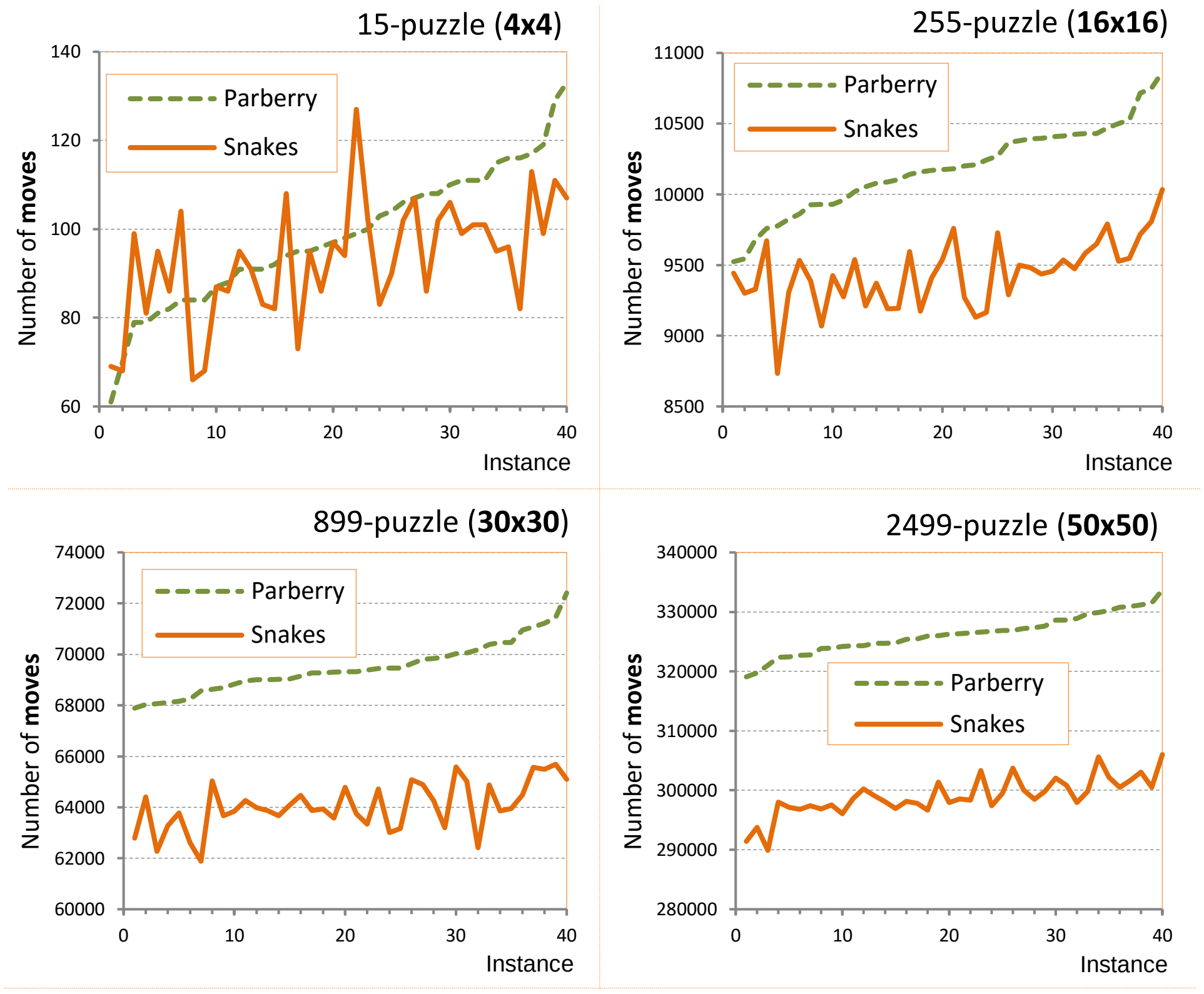

**Figure 14.** *Development of the improvement in* ***Parberry's algorithm*** *with the growing size of the puzzle instance.* Comparison of the number of moves conducted by the algorithm of *Parberry* and by its snake-based improvement is shown for four puzzles of the increasing size. Individual instances for each size of the puzzle are sorted according to the increasing number of steps made by *Parberry's* algorithm. It is observable that a worsening after applying snake-based approach may appear in small instances. Nevertheless, the improvement is becoming stable (between 8-9%) for larger instances.

## 6.2. *Experimental Evaluation of Snakes in the BIBOX Algorithm*

The *BIBOX* algorithm can be used to solve $(n^2-1)$-puzzle instances, as they are special cases of CPF. We evaluated *BIBOX* algorithm under the same set of tests as the algorithm of *Parberry*. As the *BIBOX* algorithm is more general – it solves instances of cooperative path-finding over bi-connected graphs while the underlying 4-connected grid of the $(n^2-1)$-puzzle is a special case of bi-connected graph – the *BIBOX* algorithms has been submitted to additional tests over more general bi-connected graphs.

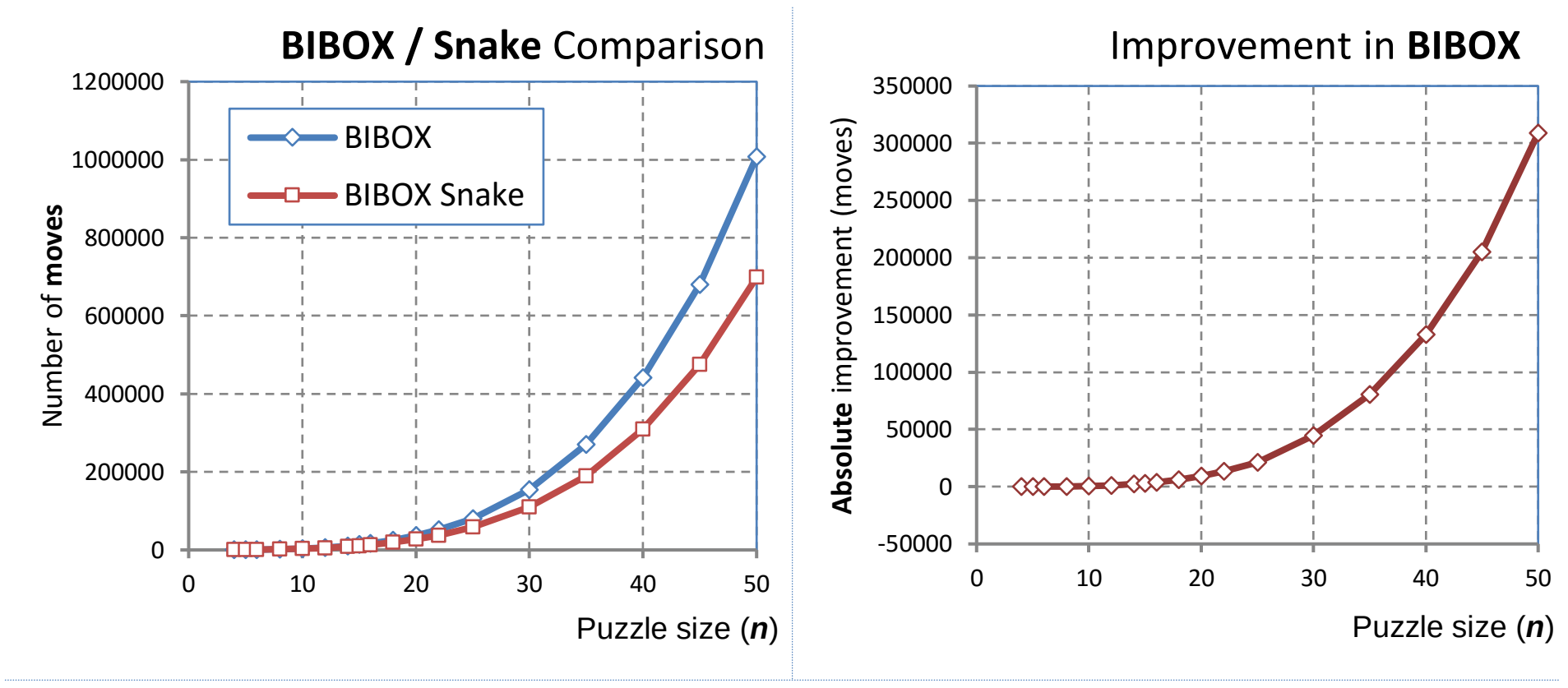

**Figure 15.** *Comparison of the original* ***BIBOX algorithm*** *and its modification that uses* ***snakes****.* The experimental setup is the same as in the case of *Parberry's* algorithm – $(4^2 - 1)$ to $(50^2 - 1)$ puzzles with 40 random instances for each size were used for evaluation. The average number of steps over 40 instances is shown for each size of the puzzle. The *BIBOX* algorithm needs approximately 3-times more steps than *Parberry's* algorithm to solve an instance of the $(n^2 - 1)$-puzzle. The absolute improvement in terms of the number of steps after introducing snake-based movements into *BIBOX* algorithm is larger in absolute number of steps than after using snakes within the Parberry algorithm.

### 6.2.1. *Competitive Comparison of the BIBOX Algorithm with Snake Improvement on Puzzle Instances*

Let us recall that the original *BIBOX* algorithm as presented in [26] requires two unoccupied vertices so it not immediately applicable to $(n^2 - 1)$-puzzle. These two unoccupied vertices are needed to arrange robots/pebbles in the initial cycle of the ear decomposition while just one vacant vertex is sufficient to arrange pebbles in regular ears [33]. There exists a variant of the algorithm called *BIBOX-$\theta$* [27, 33] that suffices with just one unoccupied vertex that uses special transposition rules to arrange pebbles in the initial cycle. This algorithm is a candidate to be used in our experiments but as we are working with graphs of fixed 4-connected structure in $(n^2 - 1)$-puzzle there is no need to use the general transposition rules of *BIBOX-$\theta$* for the initial cycle. Instead, a look-up table with optimal solution to all the possible rearrangements in the original cycle consisting of 4 vertices has been used.

There is also a possible parameterization of the *BIBOX* algorithm with the ear decomposition to be used. As it is shown in [35] various ear decompositions can affect the number of generated moves by the algorithm greatly. Hence, we fixed the ear decomposition of the 4-connected grid so that internal ears consist of a single vertex (that is, we start with the initial 4-cycle and continue by adding ears that fills the underlying 4-connected grid row by row). Such ear decomposition is most similar to how *Parberry's* algorithm proceeds.

| Relative Improvement in BIBOX Algorithm | |
|---|---|
| $n$ | **Length Improvement** (%) |
| 4 | -9.29 |
| 5 | -1.46 |
| 6 | 2.72 |
| 8 | 13.98 |
| 10 | 18.48 |
| 12 | 19.71 |
| 14 | 22.08 |
| 16 | 22.42 |
| 18 | 23.54 |
| 20 | 24.45 |
| 22 | 26.03 |
| 25 | 26.46 |
| 30 | 26.88 |
| 35 | 29.04 |
| 40 | 29.79 |
| 45 | 30.07 |
| 50 | 30.14 |

**Table 2.** ***Relative improvement*** *achieved by using snakes in the BIBOX algorithm.* Again, the improvement has been measured for several sizes of the puzzle ranging from $(4^2-1)$ to $(50^2-1)$ with 40 random instances per size. Relative improvement is significantly larger in the case of *BIBOX* algorithm than in Parberry's algorithm.

**Figure 16.** *Illustration of the* ***trend*** *in the average improvement in the BIBOX algorithm.* It can be observed that the relative improvement tends to stabilize around 30% as instances of the $(n^2-1)$-puzzle are getting larger.

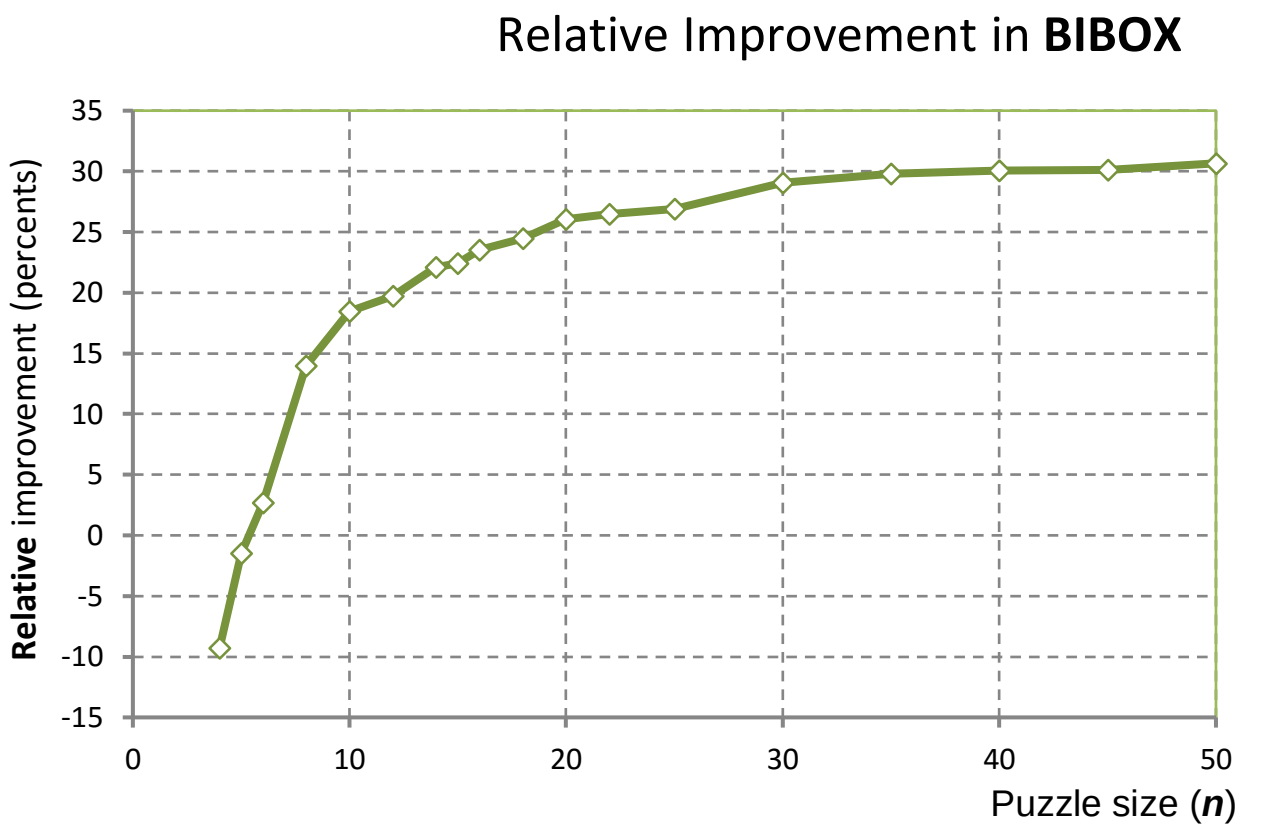

Results from the comparison of the number of steps in solutions of the puzzle generated by the *BIBOX* algorithm and its snake-based improvement are shown in Figure 15. Observe that the *BIBOX* algorithm generates approximately 3 times larger solutions than the algorithm of Parberry. This is however natural result as *BIBOX* algorithm is more general for bi-connected graphs and does not exploit the advantage of a priori knowledge that the underlying graph is a 4-connected grid. It is also noticeable that using snakes in case of the *BIBOX* algorithm saves much more steps in absolute terms than in the case of *Parberry's* algorithm.

The relative improvement after introducing snakes into the *BIBOX* algorithm as shown in Figure 16 and Table 2 is around 30% in larger puzzle instance. In small instances even worsening may appear which is caused by inaccuracy of the distance heuristic (7) which does not take into account that the number of performed moves does not need to correspond to shortest paths and that the second robot (denoted as $s$ in section 5.2) may relocate after relocating the first robot (denoted as $r$).

### 6.2.2. *BIBOX Algorithm on Individual Puzzle Instances*

Similarly as in the case of *Parberry's* algorithm, we show results of test of the *BIBOX* algorithm over individual instances of the puzzle. Results are shown in Figure 17. In small instances, relatively significant worsening may appear after using snakes in algorithm

*BIBOX*. On the other hand, in large instances significant improvement over 30% can be achieved. Again, the worsening in small instances can be explained by inaccuracy of distance heuristic (7) as in small instances stronger interference between two relocated robots is more likely.

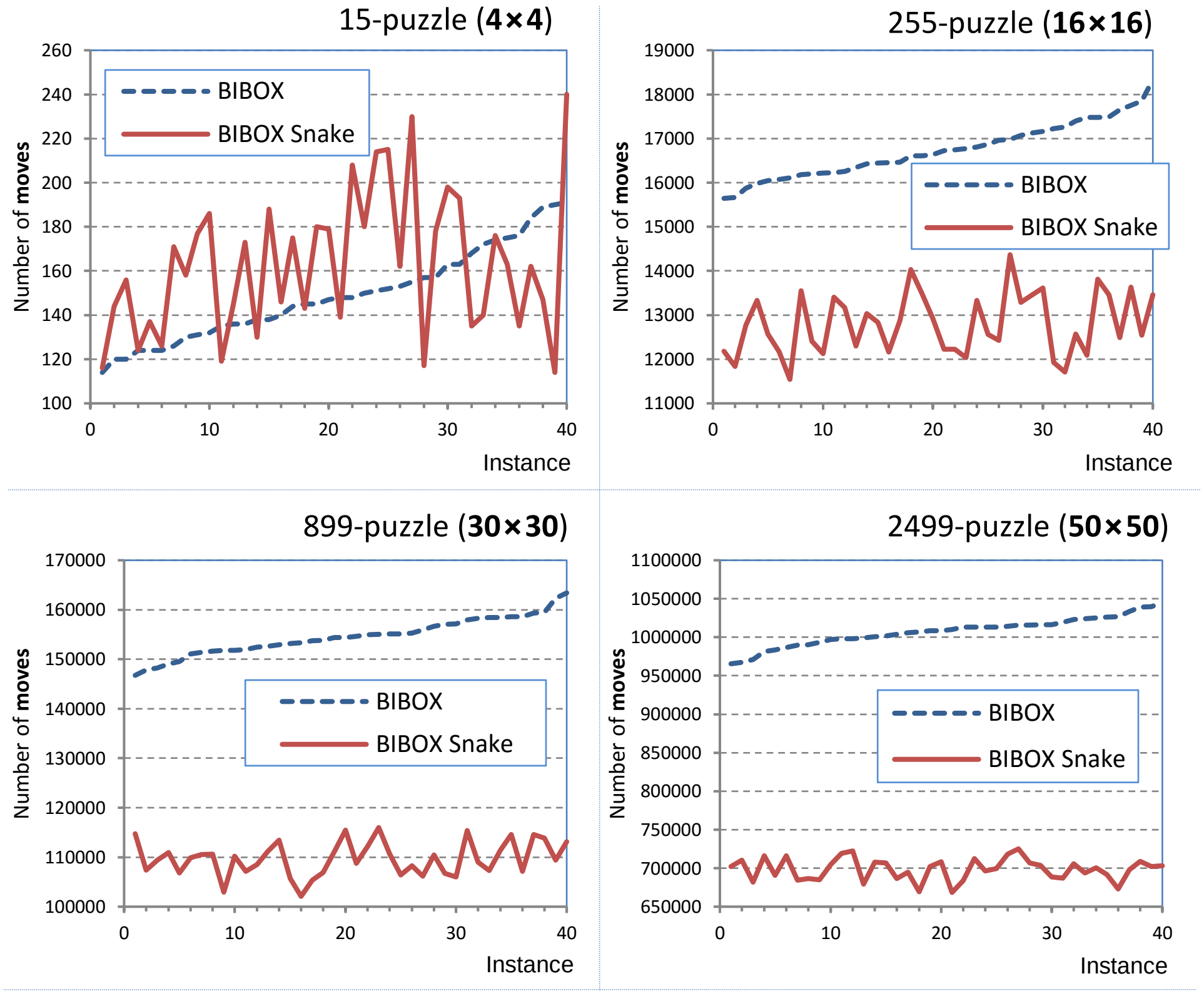

**Figure 17.** *Development of the improvement in the* ***BIBOX algorithm*** *with the growing size of the* ***puzzle*** *instance.* The improvement is shown for all the 40 random instances for several sizes of the $(n^2 - 1)$-puzzle. Worsening may appear after using snakes in the *BIBOX* algorithm in small instances – the same behavior can be observed in *Parberry's* algorithm. Nevertheless, the improvement is becoming stable around 30% in larger instances.

### 6.2.3. *Evaluation of Using Snakes in BIBOX Algorithm on CPFs over Bi-connected Graphs*

Promising results obtained in solving $(n^2 - 1)$-puzzle by snake-improved algorithms inspired us to evaluate snake-based version of the *BIBOX* algorithm on instances of CPF over more general bi-connected graphs that are structurally different from the puzzle.

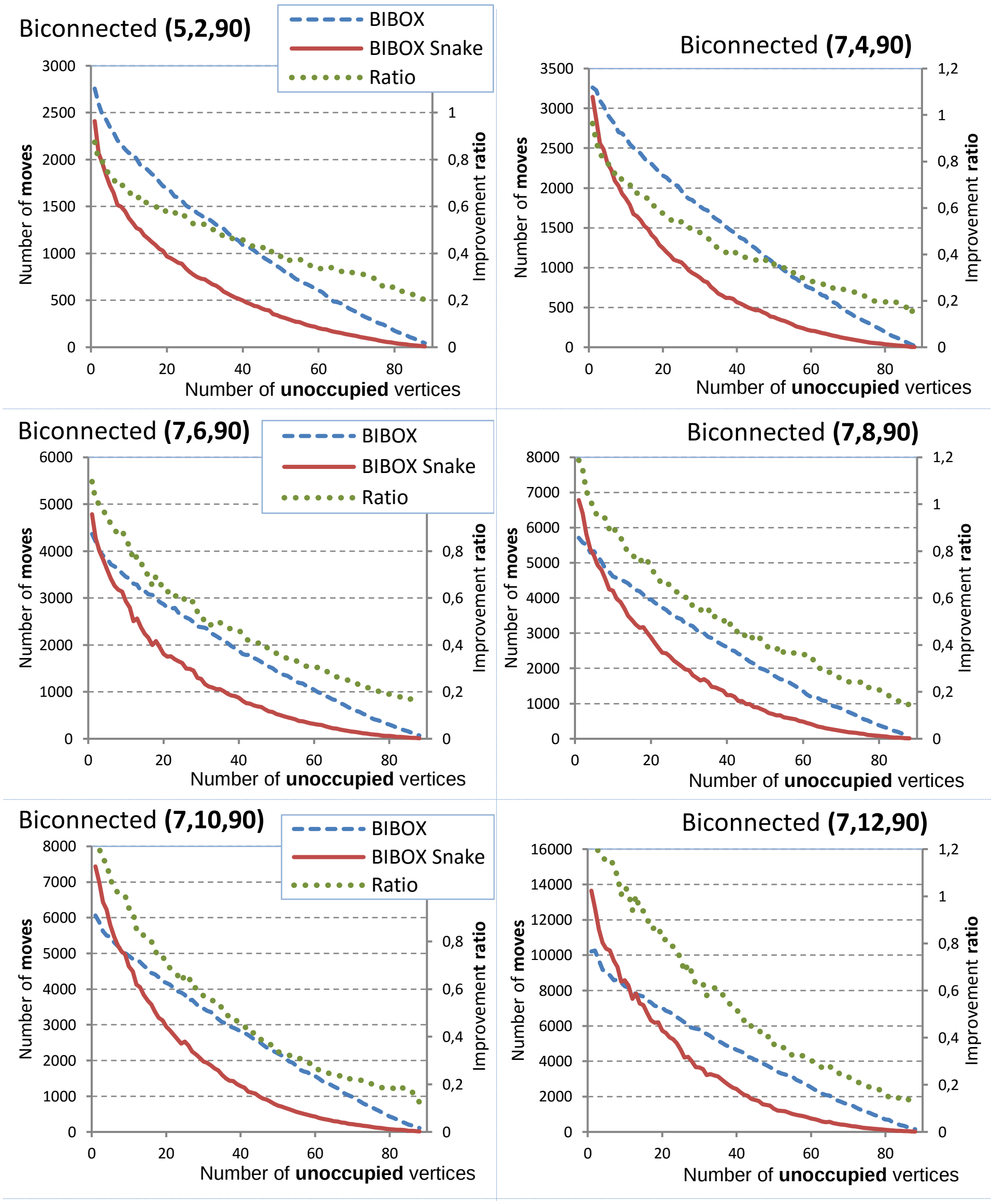

**Figure 18.** *Improvements after* ***introducing snakes into BIBOX*** *algorithm executed on random* ***bi-connected*** *graphs.* Bi-connected graphs with different sizes of ears in ear decompositions were tested – $Biconnected(c, e, |V|)$ stands for a bi-connected graph with initial cycle of size $c$, average internal ear size $e$, and the number of vertices $|V|$. The number of robots changed from almost fully occupied graph (one vacant vertex) to a graph occupied by a single robot. Again, the average number of steps in solutions to 40 random instances for each number of occupied vertices is shown. In almost all the cases, snakes bring significant improvement of the total number of steps – the original solution has been reduced by up to 80%. There is also observable tendency that snakes appear to be more beneficial if multiple vacant vertices are available. The only case, where snakes cause worsening, consists of almost fully occupied graphs with relatively long ears in the decomposition.

This is motivated by the fact that the ear decomposition of the 4-connected grid, where $(\mathrm{n}^2-1)$-puzzle takes place, is quite special – it consists of ears having just one internal vertex in most case. Hence, it would be interesting to see how snake improvement behaves in *BIBOX* algorithm over bi-connected graphs with longer ears.

In addition to structurally different underlying graphs, we evaluated snake-based improvements over CPF instances in presence of various numbers of robots; that is, when multiple vacant vertices are available. Availability of multiple unoccupied vertices may affect snake formation and interference between two robots relocated jointly significantly. We expect the higher accuracy of the distance heuristic (7) in cases of with fewer robots (more vacant vertices) as there should be weaker interference between two relocated robots.

A series of tests with CPFs over bi-connected graphs with various ear decompositions has been also done to evaluate benefits of snakes in situations structurally different from those in $(n^2-1)$-puzzle.

Several random bi-connected graphs were constructed over which random CPF instances were generated. Random instances over fixed graph are obtained by generating random initial and goal configuration of robots. Results from the evaluation on CPFs over bi-connected graphs are shown in Figure 18.

The notation $Biconnected(c, e, m)$ denotes random bi-connected graph with an ear decomposition where the initial cycle consists of $c$ vertices, the average number of internal vertices of ears is $e$, and the total number of ears is $m$. Several bi-connected graphs containing approximately 90 vertices were used in the evaluation.

Random bi-connected graph is constructed by adding ears of random length (uniform distribution where $e$ is the mean) to randomly selected endpoints in the already constructed part of the graph. The construction process starts with the initial cycle of given size $c$ and terminates after the given number of vertices $m$ is exceeded.

The occupancy by robots in the tested instances grown from one robot up to as many robots so that there were only two unoccupied vertices in the graph (the standard *BIBOX* algorithm could be used).

Results with the *BIBOX* algorithm over random bi-connected graphs indicate that the snake-based improvement is more efficient if there are more unoccupied vertices in the instance, which conforms to our expectation (though the scale of improvement was not expected). Up to $50\%$ moves can be saved after employing snakes in CPF solving over bi-connected graphs with approximately half of vertices occupied by robots and even larger proportion (up to $80\%$) of moves can be saved in sparsely occupied graphs.

Worsening after using snakes appears more frequently than in the case of puzzle. It may appear especially with long ears in densely occupied graphs. This behavior can be explained by the fact that there are fewer alternative paths in bi-connected graphs with long ears and by the fact that robot can sometimes be influenced by other robots in densely occupied graph which can divert it from its direction. Due to absence of alternative paths the diversion cannot be repaired as easily as in the case of $(n^2-1)$-puzzle – in other words distance heuristic (7) tends to be quite inaccurate in such cases.

### 6.3. ***Experimental Evaluation of Snakes in the PUSH-and-SWAP Algorithm***

The last of algorithms included in our testing and snake-based improvements is *PUSH-and-SWAP*. The algorithm is fundamentally different from other two algorithms in its non-local behavior. Let us recall that when the *Swap* or the *Twin-Swap* operations are about to be executed to exchange a pair or a triple of robots respectively, a vertex with enough neighbors is searched where the exchange can be conducted. Such a vertex with enough neighbors may be potentially located far from the current occurrence the pair or the triple of robots (especially in the case of bi-connected graphs with long ears). Hence, these operations may take place over a significant part of the environment while many robots can be affected.

Almost the same set of testing instances of the puzzle has been used for the *PUSH-and-SWAP* algorithm as in the previous tests. The algorithms however needs at least two unoccupied vertices and in its snake-based improved variant at least three unoccupied vertices are needed. Hence instead of solving the $(n^2 - 1)$-puzzle the $(n^2 - 3)$-puzzle has been solved to be able to make any comparison. The generation of instances of the $(n^2 - 3)$-puzzle differs from those for $(n^2 - 1)$-puzzle in not adding the last two pebbles.

The *PUSH-and-SWAP* algorithm is yet more general than *BIBOX* as it can solve CPF instances over arbitrary graph with at least two unoccupied vertices. However, in our comparison it was sufficient to be restricted on bi-connected graph tests where we used the same set of instances as in the case of the *BIBOX* algorithm with omitting instances containing only two unoccupied vertices.

The known problematic behavior of the *PUSH-and-SWAP* algorithm described in [42, 43] due to which the *Rotate* operation has been introduced does not occur in our testing scenarios.

#### 6.3.1. *Competitive Comparison of the PUSH-and-SWAP Algorithm with Snake Improvement*

An important parameter of the *PUSH-and-SWAP* algorithm is the linear ordering of vertices/robots, which is followed when robots are placed one by one to their goal positions. Since the snake-improvement is sensitive to locality as indicated by the previous experiments and by additional experiments conducted during the development, we used ordering of robots that preserves locality in the environment. That is, robots that are close to each other in the ordering should be close to each other in the graph. Such an ordering can be obtained by the breadth first search.

The comparison of the number of moves generated by the *PUSH-and-SWAP* algorithm and its snake improvement on $(n^2 - 3)$-puzzle instances is shown in Figure 19. It can be clearly observed that the improvement obtained by snakes is small in comparison with the size of the original solution though in absolute terms the many moves are saved. Although the absolute improvement grows for growing size of the puzzle instances, the improvement growth is not stable.

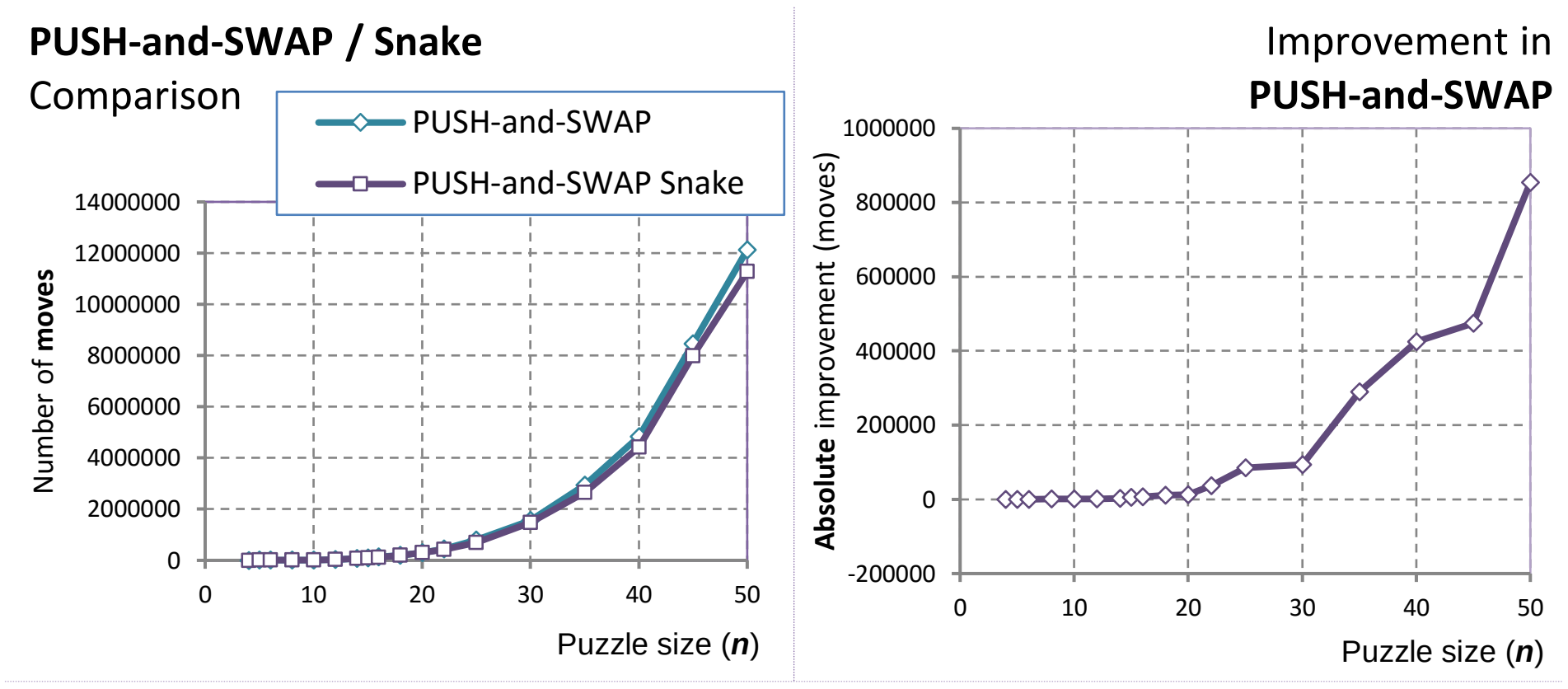

**Figure 19.** *Comparison of the original* ***PUSH-and-SWAP algorithm*** *and its* ***snake-based improvement*** *in terms of total number of steps.* Comparison has been done on the same set of instances as in the case of other tested algorithm but with two additional pebbles removes to obtain $(n^2 - 3)$-puzzle instance. The *PUSH-and-SWAP* algorithm generates up to 10 times more moves than *BIBOX* (despite it has two more free vertices) while *BIBOX* generates up to 3 times more moves than the *Parberry's* algorithm on tested puzzle instances.

Further investigation of the relative solution improvement achieved by using snakes is presented in Figure 20 and Table 3. The relative improvement is between 3% to 10% and most frequently around 6%, which renders the *PUSH-and-SWAP* algorithm to be the worst candidate for the snake improvement among the tested three algorithms for the puzzle instances. Moreover, the relative improvement is unstable for growing size of the instance. That is, having bigger instance does not necessarily imply larger relative improvement of the solution. There is also no sign of convergence of the relative improvement for growing puzzle instances.

6.3.2. *PUSH-and-SWAP Algorithm on Individual Puzzle Instances*

The hypothesis explaining the unstable relative improvements in the solutions is that local improvement – that is, the joint relocation of a pair of robots – may lead to a global worsening of the solution. This is quite likely in the case of *PUSH-and-SWAP* algorithm since the relocation of a pair of robots often requires extensive swap of a triple of robots (*Twin-Swap* operation) which affects many other robots except the swapped triple and makes it hardly predictable if the local improvement propagates to the global quality of the solution.

The behavior of the snake improvement on individual puzzle instances is presented in Figure 21. As in the previous tests, we evaluated the algorithm and its snake improvement on 40 random instances generated for each size of the puzzle – $4\times4$, $16\times16$, $30\times30$, and $50\times50$. The instances in the collection of 40 instances for each size of the puzzle are sorted according to the growing number of moves generated by the original *PUSH-and-SWAP*.

Results confirm the hypothesis that the local improvement obtained by using snakes often leads to the worsening of the size of the overall solution. Worsening appears more frequently when original PUSH-and-SWAP generates relatively small solution.

| **Relative Improvement** in PUSH-and-SWAP Algorithm | |
|---|---|
| $n$ | **Length Improvement** (%) |
| 4 | 3,31 |
| 5 | 4,42 |
| 6 | -0,27 |
| 8 | 6,03 |
| 10 | 5,94 |
| 12 | 3,81 |
| 14 | 3,17 |
| 16 | 4,74 |
| 18 | 5,44 |
| 20 | 5,85 |
| 22 | 4,36 |
| 25 | 8,10 |
| 30 | 10,75 |
| 35 | 6,02 |
| 40 | 9,90 |
| 45 | 8,78 |
| 50 | 5,61 |

**Table 3.** ***Relative improvement*** *achieved by using snakes in the PUSH-and-SWAP algorithm.* Again, the same set of testing instances of the $(n^2 - 3)$-puzzle has been used. The improvement is smaller in the *PUSH-and-SWAP* algorithm than in other two tested algorithms.

**Figure 20.** *Illustration of the* ***trend*** *in the average improvement in the PUSH-and-SWAP algorithm.* It can be observed, that the relative improvement is visibly unstable with respect to the increasing size of the instance.

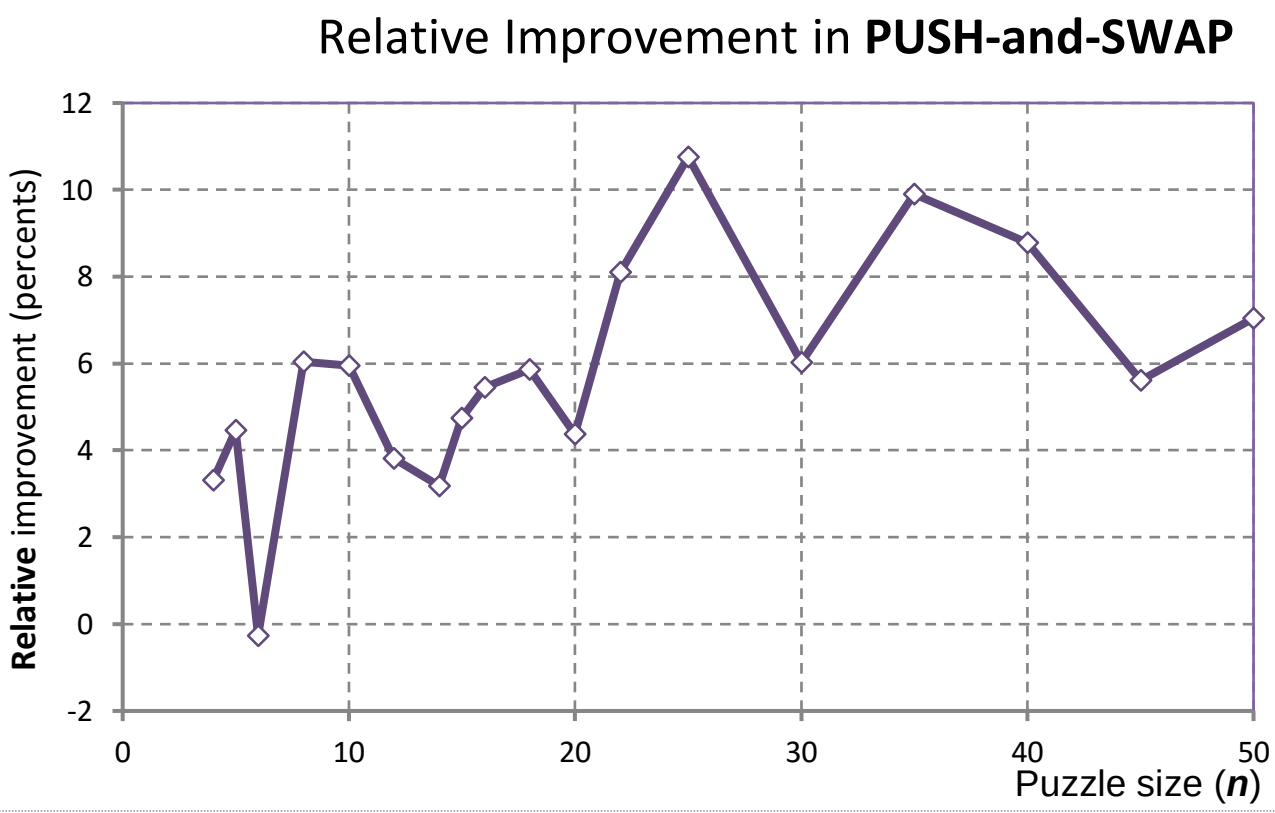

6.3.3. *Evaluation of Using Snakes in the PUSH-and-SWAP Algorithm on CPFs over Bi-connected Graphs*

Finally, we made experimental evaluation of the *PUSH-and-SWAP* algorithm on random bi-connected graphs. Again, the same instances have been used as for other two algorithms. Results for random bi-connected graphs are presented in Figure 22.

Results indicate again relatively unstable improvements between 5% and 10% by using snakes in the *PUSH-and-SWAP* algorithm. However, worsening by up to 10% appears frequently as well. It is observable that the relative improvement becomes better in instances with longer ears in the ear decomposition while in instance with short ears significant worsening by more than 10% appears frequently.

The slightly better behavior for longer ears can be explained by the cheaper relocation of a pair of robots jointly along long ears than their separate relocation. When the swapping is needed during the relocation of a pair or triple of robots, the affected part of the graph traversed during the *Swap* operation or *Twin-Swap* respectively by visiting a distant vertex with enough number of neighbors is not much different. Hence, both operations produce approximately the same number of moves while in the case of the separate relocation of the pair of agents all the moves needed to reach the distant vertex with enough neighbors where exchange of robots is done need to be executed twice. Thus the benefit of the joint relocation is more expectable with longer ears as the necessary vertex with many neighbors is more likely to be far from the current location of the pair or triple of robots being swapped.

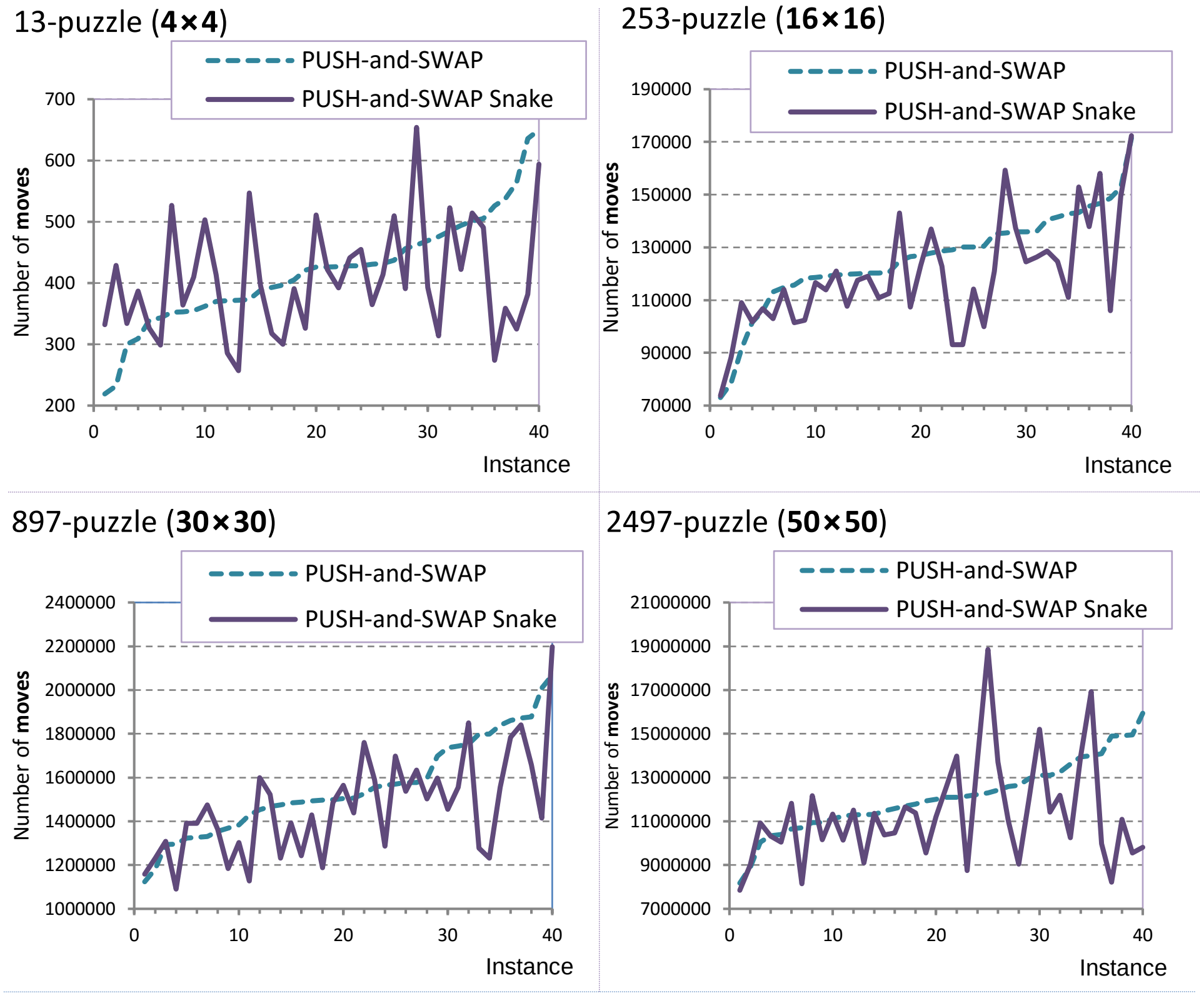

**Figure 21.** *Development of the improvement in the* ***PUSH-and-SWAP algorithm*** *with the growing size of the* $(\boldsymbol{n^2} - \mathbf{3})$-***puzzle*** *instance.* The improvement is shown for the same set of 40 instances for each size of the puzzle as in the previous tests.

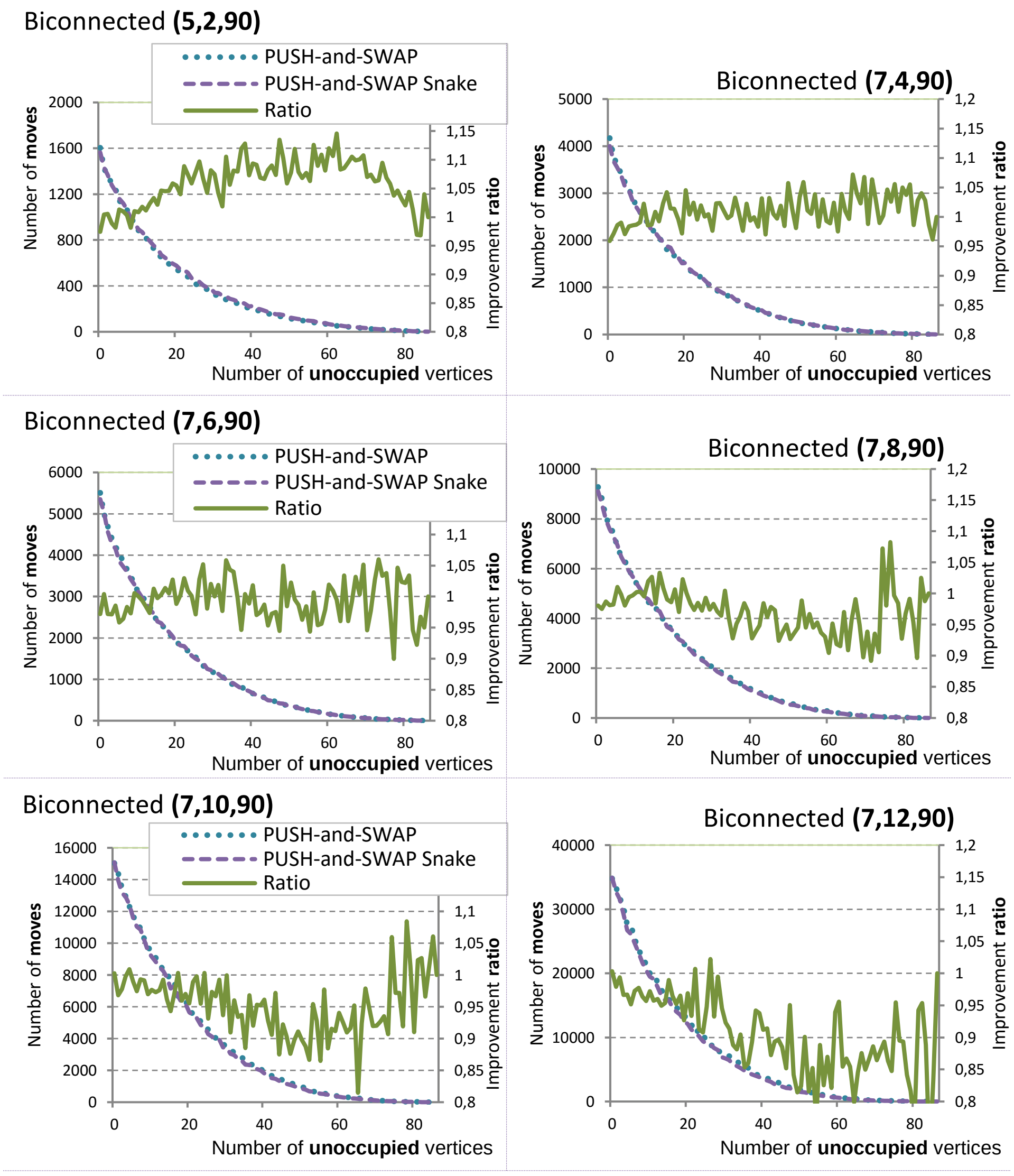

**Figure 22.** *Improvements achieved by using* ***snakes in the PUSH-and-SWAP algorithm*** *executed on random* ***bi-connected*** *graphs.* The same collection of random bi-connected graphs as in the testing of *BIBOX* algorithm (Figure 18) has been used. The improvement in the *PUSH-and-SWAP* algorithm is much smaller than in the case of *BIBOX* – only the reduction by up to 5%-10% has been achieved. In case of bi-connected graphs with short ears, the application of snakes causes marginal worsening by up to 10%. As the size of ears in the ear decomposition grows, the improvement with snakes becomes more significant. It is also observable that the improvement achieved by the snake application in *PUSH-and-SWAP* algorithm is unstable in comparison with that in the case of the *BIBOX* algorithm. The side-result readable from Figure 18 and Figure 22 is that *PUSH-and-SWAP* produces shorter solutions on bi-connected graphs with short ears while *BIBOX* produces shorter solutions on bi-connected graphs with long ears.

## 6.4. *Runtime Measurement*

Finally, results regarding runtime are presented in Table 4. The average runtime for puzzles of size up to $50\times50$ are shown (in case of the *PUSH-and-SWAP* algorithm it is $(n^2-3)$-puzzle and $(n^2-1)$-puzzle in other cases).

Expectably, snake-based improvement of *Parberry's* algorithm is slower as it makes decisions that are more complex (in fact, it is running the original algorithm plus snake placement to compare if snake is locally better). Nevertheless, the slowdown is well acceptable.

Both algorithms – *Parberry's* and its snake-based improvement – are capable of solving puzzles with solutions consisting of hundreds of thousands of moves almost immediately. Hence, it can be concluded that both algorithms scales up extremely well and they can be used in on-line applications (the scalability is indeed not because of the use of snakes – this is due to scalability of the original algorithm).

The absolute time in the case of the *BIBOX* algorithm is much worse since the algorithm works on general bi-connected graph while the *Parberry's* algorithm works on fixed grid thus there is much less decisions in the *Parberry's* algorithm.

The important result is the difference between the original and snake-improved version in case of the *BIBOX* algorithm. It is noticeable that although snakes require more complex computations these in fact should not increase the runtime significantly – the distance heuristic (7); that is, the distance between currently placed robot and the next robot to be placed can be calculated by looking into table containing all-pairs of shortest paths. The time needed for this pre-calculation is dominated by the runtime of the rest of the *BIBOX* algorithm theoretically as well as empirically. The performance in terms of the runtime is better when snakes are utilized because the algorithm does need to produce significantly fewer moves.

**Table 4.** ***Runtime*[1] *measurements*** *of the algorithm of Parberry, the BIBOX algorithm, the PUSH-and-SWAP algorithm and their snake variants.* Average time is calculated for each size of the puzzle out of 40 runs with different random setups. It can be observed that former two algorithms scale up well while snake improvement in the *PUSH-and-SWAP* algorithm tends to be computationally expensive.

| | $n$ | 10 | 20 | 30 | 40 | 50 |
|---|---|---|---|---|---|---|
| **Time** (seconds) | **Parberry** | < 0.10 | < 0.10 | < 0.10 | 0.10 | 0.10 |
| | **Parberry/Snakes** | < 0.10 | < 0.10 | < 0.10 | 0.10 | 0.19 |
| | **BIBOX** | < 0.10 | 3.52 | 37.39 | 211.95 | 835.06 |
| | **BIBOX/Snakes** | < 0.10 | 2.99 | 30.25 | 168.50 | 657.57 |
| | **Push&Swap** | <0.10 | 1.33 | 9.85 | 40.65 | 131.41 |
| | **Push&Swap/Snakes** | 0.26 | 15.86 | 189.12 | 1121.35 | 4423.36 |

[1] All the tests with Parberry's algorithm were run on a commodity PC with CPU Intel Core2 Duo 3.00 GHz and 2 GB of RAM under Windows XP 32-bit edition. The C++ code was compiled with Microsoft Visual Studio 2008 C++ compiler. Tests with the *BIBOX* algorithm and the *PUSH-and-SWAP* algorithm were run on an experimental server with the 4-core CPU Xeon 2.0GHz and 12GB RAM under Linux kernel 3.5.0-48.

Interesting results were obtained for the *PUSH-and-SWAP* algorithm. Although *PUSH-and-SWAP* generates several times longer solutions than *BIBOX* its original version is up to several times faster than *BIBOX*. However, the snake-based improvement within *PUSH-and-SWAP* slows down the algorithm by at least one order of magnitude making it the slowest algorithm out of our portfolio. Marginal improvements obtainable by introducing snakes in *PUSH-and-SWAP* at relatively high computational cost make it an unpromising option.

## 6.5. *Summary of Experimental Evaluation*

The conducted experimental evaluation clearly shows the hat snake-based reasoning integrated to the original algorithm of *Parberry* as well as to the *BIBOX* algorithm brings significant improvements in terms of the quality of generated solutions (defined as total number of moves). This claim is experimentally supported in both $(n^2 - 1)$-puzzle and yet more distinctively in cooperative path finding instances on bi-connected graphs. The integration of the snake-based reasoning in the *PUSH-and-SWAP* algorithm did not bring as promising results as in case of other two algorithms. Although the snake-based reasoning can slightly improve the solution generated by the *PUSH-and-SWAP* algorithm this is computationally costly and hence does not represent a good trade-off.

Experiments support the claim that using snakes greedily (that is, if they are locally better) leads to global improvement of the solution even though the current configuration may be worsened sometimes from the global point of view in case of algorithms which exhibit local character of robot/pebble relocation – that is, algorithms of *Parberry* and *BIBOX*. As instances are getting larger, the improvement tends to stabilize itself between $8\%$ and $9\%$ in average in case of *Parberry's* algorithm and around $30\%$ in the case of *BIBOX* algorithm on $(n^2 - 1)$-puzzle. On larger instances – that is larger than $30\times30$ – possible fluctuations towards worsening the solution are eliminated, hence using snakes expectably leads to mentioned improvement on an individual instance.

The similar claim cannot be extended on the *PUSH-and-SWAP* algorithm as it has been shown by the conducted experimental evaluation. Local improvements by using the joint relocation of pairs of robots in the *PUSH-and-SWAP* algorithm often lead to global worsening of the solution. The explanation of this behavior is that the algorithm does not have a local behavior when robots are relocated – large parts of the graph may be affected during relocation of robots in the *PUSH-and-SWAP* algorithm, which makes it difficult to predict the impact of local improvement.

Runtime measurements show that original *Parberry's* algorithm and its snake-based improvement solve instances of tested sizes in less than $0.2s$. Thus, it can be concluded that scalability is extremely good.

Surprisingly, snake-based reasoning within the *BIBOX* algorithm improves solutions quite dramatically in CPFs on bi-connected graphs with longer ears in the decomposition and fewer robots in the graph. In such cases, snakes help the *BIBOX* algorithm to reduce the size of the solution by up to $50\%$ or even $80\%$ in sparsely populated instances.

It is also an interesting result that the *PUSH-and-SWAP* algorithm generates up to 10 times more moves than *BIBOX* (despite *PUSH-and-SWAP* has two more free vertices) while *BIBOX* generates up to 3 times more moves than the *Parberry's* algorithm on tested puzzle instances.

## 7. Conclusions and Future Work

We have presented an improvement to the polynomial-time algorithm for solving the $(n^2 - 1)$-puzzle in an on-line mode sub-optimally. The improvement is based on an idea to move pebbles jointly in groups called *snakes*, which was supposed to reduce the total number of moves. The experimental evaluation eventually confirmed this claim and showed that the new algorithm outperforms the original algorithm of *Parberry* [13] by 8% to 9% in terms of the average length of the solution. Theoretical upper bounds on the worst-case length of the solution are also better for the new algorithm as we have shown.

Regarding the runtime, the new algorithm is marginally slower due to its more complex computations, however this is acceptable for any real-life application as the runtime is linear in the number of produced moves (approximately $10^6$ moves can be produced per second).

Promising results with snake-based joint pebble moving in $(n^2 - 1)$-puzzle led us to the idea to try to integrate snake-based movement into methods for solving the problem of *cooperative path-finding* (CPF) of which the $(n^2 - 1)$-puzzle is a special case. We have integrated snake reasoning into the *BIBOX* algorithm [33] and into the *PUSH-and-SWAP* algorithm. Both algorithms operate in a similar way to the algorithm of *Parberry* (that is, robots are placed one by one and after the robot is placed it does not move any more or is restricted to move).

Improvements gained after integrating snake-based reasoning into *BIBOX* algorithm were even more significant than in case of Parberry's algorithm. Up to 30% improvement was reached in solving $(n^2 - 1)$-puzzle with algorithm *BIBOX* and up to 50% improvement has been reached in CPFs over bi-connected graphs with long ears and multiple unoccupied vertices. Moreover, the improvement in CPFs on bi-connected graphs has the growing tendency as the number of unoccupied vertices increases.

Snake based joint relocation of robots in the *PUSH-and-SWAP* algorithm improved solutions by 5% to 8% - the least of all the tested algorithms. Moreover, the improvement in *PUSH-and-SWAP* is unstable. That is, unlike in other two algorithms the absolute improvement (the number of saved moves with respect to the original version of the algorithm) does not grow steadily with respect to the growing size of instances.

The better behavior of the snake-based joint relocation of robots/pebbles is expectable in algorithms where relocation of a robot/pebble does not affect too many vertices of the underlying graph – the case of *Parberry* and *BIBOX*. On the other hand, the relocation of a robot in *PUSH-and-SWAP* algorithm may affect large portion of the underlying graph as suitable vertex with enough neighbors need to be reached many times which may significantly distort local improvements of the snake based relocation in the global outcome.

It will be interesting for future work to add more measures for reducing the total number of moves towards the optimum. Choosing a more promising local rearrangement among several options can be easily parallelized.

We are also interested in generalized variants of the $(n^2 - 1)$-puzzle where there is more than one vacant position. These variants are known as the $(n^2 - k)$-puzzle with $k > 1$ [29]. Although it seems that obtaining optimal solutions remains hard in this case, multiple vacant positions can be used to rearrange pebbles more efficiently in the sub-optimal approach.

It seems that adapting the *BIBOX* algorithm for snakes of length more than 2 is also possible. A robot can collect the snake along its relocation towards the ear connection vertex. Long snakes however bring significant technical difficulties as it may happen that the snake intersects itself – an opportunistic formation of a snake and eventual break-up of the snake when it is not longer maintainable is a possible option.

Another open question is how the snake-based approach could perform in the *directed version* of CPF [1, 45]. Unidirectional environment puts additional constraints on relocation and hence solution reduction using snakes may have greater effect.

Finally, it is interesting for us to study techniques for optimal solving of this and related problems; especially the case with small unoccupied space (that is, with $k \ll n^2$). This is quite open area as today's optimal solving techniques [25] can manage only small number of pebbles compared to the size of the unoccupied space.

### Acknowledgments

This is supported by Charles University in Prague within the PRVOUK and UNCE projects. We would like gratefully thank reviewers for their comments, which significantly helped us to improve the paper and gave us valuable advice.

## Appendix A – Analysis of the Average Case

Regarding the average case analysis we will assume in accordance with [13] that every initial configuration of pebbles can occur with the same probability[1]. We will first show that algorithm of Parberry [13] produces $4n^3 - \frac{1}{2}n^2 + \frac{3}{2}n - 70$ moves in the average case. Then we will simulate this analysis also for our snake-based algorithm. Unfortunately, it seems not to be possible to express the average number of moves as any simple formula in the case of the snake-based algorithms. However, we can provide some arguments that the average solution length of the snake-based algorithm is strictly better than that of Parberry's algorithm.

Before we start with proofs of main propositions, we will introduce several technical lemmas. Proofs of these lemmas are omitted since they are easy and rather technical (detailed proofs can be found in [12]).

**Lemma 4.** The average value of $\mu_n((1,1);(x,y))$ for $x, y \in \{1,2,\ldots,n\}$ (that is the average Manhattan distance from the position $(1,1)$) is $n-1$. ■

**Lemma 5.** The average value of $\mu_n((1,k);(x,y))$ for $k, x, y \in \{1,2,\ldots,n\}$ such that for $x > k$ or for $y > 1$ (that is, for a given $k$ we consider only $(x,y)$ positions that follows the position $(1,k)$ in the top-down/left-right direction) is at most $n - \frac{1}{2}$. ■

The similar result can be obtained for positions in the first column. But here the estimation of the Manhattan distance is lower – namely $n-1$.

**Lemma 6.** The number of moves necessary to move a pebble from a position $(i,j)$ to a position $(1,k)$ supposed that the position $(1,k)$ is unoccupied is at most $6\mu_n\big((1,k);(i,j)\big) + 1$. ■

[1] Results presented in this appendix were superseded by the recent works of Parberry [14, 15] where better average case bounds of have been shown. These results were not known at the time of submission of this paper hence we decided to move the average case analysis from the main text to the appendix.

***Proposition 3 (Average-case Solution Length - Parberry).*** The average length of solutions to $(n^2 - 1)$-puzzle produced by Parberry's algorithm is at most $4n^3 - \frac{1}{2}n^2 + \frac{3}{2}n - 70$. ■

**Proof sketch.** From the Lemma 5 and Lemma 6 we can obtain that the expected number of moves necessary to solve the first row of the puzzle is at most: $n \cdot \left(6 \cdot \left(n - \frac{1}{2}\right) + 1\right) = 6n^2 - 2n$. Similarly for the first column: $(n-1) \cdot (6 \cdot (n-1) + 1) = 6n^2 - 11n + 5$. Altogether the upper estimation of the number of moves to solve the first row and the first column is: $12n^2 - 13n + 5$.

Suppose that the position $(1,1)$ is unoccupied and let us denote $S(n)$ the estimation of the number of moves to solve the entire $(n^2 - 1)$-puzzle. Then it holds that: $S(n) = S(n-1) + 12n^2 - 13n + 5$, where $S(3) = 34$ (calculated as the upper for average length of optimal solutions). After solving the recurrent equation we obtain that: $S(n) = 4n^3 - \frac{1}{2}n^2 + \frac{3}{2}n - 70$. ■

Notice, that this is a new theoretical result for the Parberry's algorithm (in [13] only the worst case upper bound of $5n^3 + O(n^2)$ and lower bounds are given).

**Observation 1 *(Average-case Solution Length – Snake-based).*** The average length of solutions to $(n^2 - 1)$-puzzle generated by the Snake-based algorithm is strictly lower than that of solutions generated by the algorithm of Parberry. ■

**Sketch of proof.** The average length of the solution in random instances in the case of the snake-based algorithm can be expressed as the average number of moves necessary to place first two pebbles (top-down/left-right direction) plus the average solution length to instances where first two pebbles are already placed. Notice that the average number of moves to place the first two pebbles is strictly lower in the snake-based algorithm. Hence, if we unfold the recurrence expression for the average length of the solution entirely the result will be strictly smaller than the average length of solutions of Parberry's algorithm. ■

Although we don't provide any explicit formula for the average length of solutions generated by our snake-base algorithm we know that it is strictly less than $4n^3 - \frac{1}{2}n^2 + \frac{3}{2}n - 70$.

### Appendix B – BIBOX Algorithm with Snakes

A commented pseudo-code of the *BIBOX* algorithm enhanced with snakes is given in this appendix. The original *BIBOX* algorithm arranges robots into ears while the problem inductively reduces on a smaller bi-connected graph whenever robots are arranged into the ear – robots in such an ear do not move any more. The algorithm is listed below as Algorithm 1.

The algorithm uses several auxiliary functions to solve subtasks. Pseudo-code of auxiliary functions is given in [33] – here they are only briefly described.

The algorithm starts with constructing ear decomposition (line 1). It is assumed that a cycle denoted as $C(L_i)$ is associated with each ear; $C(L_i)$ can be constructed by adding a path connecting ear's connection vertices $u$ and $v$. Then the goal configuration of robots is transformed so that vacant vertices are eventually located in the initial cycle of the decomposition (line 2). The algorithm solves this modified instance afterwards. The solution of the original instance is obtained by relocating vacant vertices from initial cycle to their original goal locations (line 8). This instance transformation is carried out by auxiliary functions *Transform-Goal* and *Finish-Solution* that relocates vacant vertices along two vertex disjoint paths. The main loop (lines 4-6) processes ear from the last one towards the initial cycle. Robots are arranged by another auxiliary procedure *Solve-Original-Cycle* in the original cycle (line 7).

Individual ears are processed by the procedure *Solve-Regular-Ear*. It arranges robots into the ear in stack like manner. First, unoccupied vertices are moved out of the processed ear as they will be needed there (lines 10-14). Then robots, whose goal positions are in the ear, are processed. Two cases are distinguished depending on whether the processed robot is located outside the ear (lines 17-25) or within the ear (lines 27-51).

The easier case is with robot outside – in this case, the robot is moved to the connection vertex $u$ using either *Move-Robot* or *Move-Robot-Snake* auxiliary procedure. The other connection vertex $v$ is vacated by *Make-Unoccupied* procedure. If some vertex is free on the cycle $C(L_c)$ then the cycle can be rotated which is done once in the positive direction by *Rotate-Cycle*$^+$ function. The rotation places the robot into the ear. Throughout the relocation of robots vertex locking is used (functions *Lock* and *Unlock*) to fix an robot in certain vertex while other robots or vacant vertex are relocated.

A more difficult case appears if the robot is inside the handle. In such case, the robot must be rotated out of the handle to the rest of the graph (lines 30-32). The number of positive rotations to get the robot out of the handle is counted (lines 27-32). The counted number of rotations is used to restore the situation by the corresponding number of negative rotations (lines 42 -44). At this point, the situation is the same as in the previous case. Thus, the robot is stacked into the handle in the same way.

The difference of *BIBOX* algorithm with snakes from the original *BIBOX* algorithm consists in adding *Move-Robot-Snake* procedure. When robots are relocated towards the currently processed ear, the snake based reasoning considers two consecutive robots whenever possible (lines 18-19 and 35-36). That is, while in the original algorithm, a single robot has been always relocated, in the snake version, the next to be relocated robot is considered as well. If both consecutive robots are close enough to each other they are relocated towards their target ear together jointly (the process of the joint relocation is implemented within *Move-Robot-Snake* procedure).

**Algorithm 3.** *The **BIBOX with snakes** algorithm.* The algorithm solves cooperative path-finding problem (CPF) over bi-connected graphs consisting of a cycle and at least one ear with two unoccupied vertices. The algorithm proceeds inductively according to the ear decomposition. The two unoccupied vertices are necessary for arranging robots within the initial cycle in the rest of the graph only one unoccupied vertex is needed. The pseudo-code is built around several higher-level operations. The modification from the original version consists in placing robots into an ear where two consecutive robots are considered at once. If consecutive robots are close enough to each other they are relocated towards the ear in a snake like manner together.

- Lock($U$) — locks all the vertices from set $U$; each vertex is either *locked* or *unlocked*; an robot must not be moved out of the locked vertex which is respected by other operations
- Unlock($U$) — unlocks all the vertices from set $U$
- Make-Unoccupied($v$) — vacates vertex $v$ sensitively to locked vertices
- Make-Unoccupied’($v$) — vacates vertex $v$ irrespective of locked vertices
- Move-Robot($r, v$) — moves robot $r$ from its current location to vertex $v$
- Move-Robot-Snake($r, s, v$) — moves robots $r$ and $s$ from their current locations towards $v$; that is $r$ is moved to $v$ and $s$ is moved together with $s$ in a snake-like manner if $r$ and $s$ are close enough initially ($\text{dist}_G(v, \alpha(r)) + \text{dist}_G(\alpha(r), \alpha(s)) < \text{dist}_G(v, \alpha(r)) + \text{dist}_G(v, \alpha(s))$; that is, the total distance towards destination $v$ is smaller if robots go together than if they go one by one)
- Rotate-Cycle$^+$($C$) — rotates cycle $C$ in the positive direction; a vacant vertex must be present in the cycle
- Rotate-Cycle$^-$($C$) — rotates cycle $C$ in the negative direction
- Transform-Goal($G, R, \alpha_+$) — transforms goal configuration $\alpha_+$ to a new configuration so that finally unoccupied vertices are located in the initial cycle of the ear decomposition; two disjoint paths along which empty vertices are relocated are returned
- Finish-Solution($\varphi, \chi$) — transforms the configuration with two unoccupied vertices in the initial cycle to the original goal configuration; $\varphi$ and $\chi$ are two disjoint paths along which empty vertices shifted
- Solve-Original-Cycle — arranges robots within the initial cycle of the ear decomposition to comply with the transformed goal configuration; two empty vertices are employed to arrange robots

---

**procedure** *BIBOX-Snake-Solve*($G = (V, E), R, \alpha_0, \alpha^+$)

/* Top level function of the BIBOX algorithm with snakes; solves a given CPF.
Parameters: $G$ - a graph modeling the environment,
$R$ - a set of robots,
$\alpha_0$ - a initial configuration of robots,
$\alpha_+$ - a goal configuration of robots. */

**let** $\mathcal{D} = [C_0, L_1, L_2, \dots, L_k]$ be a ear decomposition of $G$
$(\alpha_+, \varphi, \chi) \leftarrow$ *Transform-Goal*$(G, R, \alpha_+)$
$\alpha \leftarrow \alpha_0$
**for** $c = k, k-1, \dots, 1$ **do**
    **if** $|L_c| > 2$ **then**
        *Solve-Regular-Ear*$(c)$
*Solve-Original-Cycle*()
*Finish-Solution*$(\varphi, \chi)$

**procedure** *Snake-Solve-Regular-Ear*$(c)$
/* Places robots which destinations are within a ear $L_c$; robots placed in the ear $L_c$ are finally locked so they cannot move any more.
Parameters: $c$ – index of a ear */
**let** $[u, w_1, w_2, \dots, w_l, v] = L_c$
/* Both unoccupied vertices must be located outside the currently solved ear. */
**let** $x, z \in V \setminus \bigcup_{c=j}^{k} (L_j \setminus \{u, v\})$ such that $x \neq z$
*Make-Unoccupied*$(x)$
Lock$(\{x\})$
*Make-Unoccupied*$(z)$
Unlock$(\{x\})$
**for** $i = l, l-1, \dots, 1$ **do**
    *Lock*$(L_c \setminus \{u, v\})$
    /* An robot to be placed is outside the ear $L_c$. */
    **if** $\alpha(\alpha_+^{-1}(w_i)) \notin (L_c \setminus \{u, v\})$ **then**
        **if** $i > 1$ **then**
            *Move-Robot-Snake*$(\alpha_+^{-1}(w_i), \alpha_+^{-1}(w_{i-1}), u)$
        **else**
            *Move-Robot*$(\alpha_+^{-1}(w_i), u)$
        (**modification** w.r.t. original *BIBOX*)
        Lock$(\{u\})$
        *Make-Unoccupied*$(v)$
        Unlock$(L_c)$
        *Rotate-Cycle*$^+(C(L_c))$
    /* An robot to be placed is inside the ear $L_c$. */
    **else**
        *Make-Unoccupied*$(u)$
        Unlock$(L_c)$
        $\rho \leftarrow 0$
        **while** $\alpha(\alpha_+^{-1}(w_i)) \neq v$ **do**
            *Rotate-Cycle*$^+(C(L_c))$
            $\rho \leftarrow \rho + 1$
        Lock$(L_c \setminus \{u, v\})$
        **let** $y \in V \setminus (\bigcup_{j=c+1}^{d} (L_j \setminus \{u, v\}) \cup C(L_j))$
        **if** $i > 1$ **then**
            *Move-Robot-Snake*$(\alpha_+^{-1}(w_i), \alpha_+^{-1}(w_{i-1}), y)$
        **else**
            *Move-Robot* $(\alpha_+^{-1}(w_i), y)$
        (**modification** w.r.t. original *BIBOX*)
        Lock $(\{y\})$
        *Make-Unoccupied* $(u)$
        Unlock $(L_c)$
        **while** $\rho > 0$ **do**
            *Rotate-Cycle*$^-(C(L_c))$
            $\rho \leftarrow \rho - 1$
        Unlock $(\{y\})$
        Lock $(L_c \setminus \{u, v\})$
        *Move-Robot* $(\alpha_+^{-1}(w_i), u)$

```
|   |   Lock ({u})
|   |   Make-Unoccupied (v)
|   |   Unlock (L_c)
|_  |_  Rotate-Cycle+(C(L_c))
Lock (L_c \ {u, v})
```

## Appendix C – Simplified *PUSH-and-SWAP* Algorithm Enhanced with Snakes

A commented pseudo-code of the simplified *PUSH-and-SWAP* algorithm enhanced with snakes is given in this appendix. The original *PUSH-and-SWAP* algorithm [11] is designed to solve CPF over arbitrary graph containing at least two unoccupied vertices. Our simplification omits several special cases that ensure completeness of the algorithm over arbitrary graphs. These cases did not occur in the puzzle and random bi-connected graph instances used in the presented experimental evaluation. Moreover, the simplification also has no impact on the joint relocation of robots in pairs.

Though the *PUSH-and-SWAP* algorithm has been further corrected in [42, 43] (the corrected algorithm is called *PUSH-and-ROTATE*) we follow the original one as the cases treated in the corrected version are omitted in our simplification.

The algorithm distinguishes vertices between locked and unlocked ones similarly as in the case of the *BIBOX* algorithm. Another similarity with the *BIBOX* algorithm is that robots are placed to their destinations one by one. Whenever a robot reaches its goal vertex the goal vertex becomes locked. Robots can freely move in and out unlocked vertices but robots in locked vertices cannot move regularly; they can be temporarily moved out of their goal vertex provided they return.

The snake enhancement of the *PUSH-and-SWAP* algorithm presented here adds two major operations *Twin-Push* and *Twin-Swap* and several supporting operations. These operations represent an analogy to *Push* and *Swap* operations of the original algorithm. The two new operations move a pair of robots jointly towards the goal of the former one. In addition to *Twin-Push* and *Twin-Swap*, operations *Twin-Multipush*, *Twin-Exchange*, and few others are introduced.

The main loop of the simplified algorithm (lines 2-8) places robots one by one to their goals. If more than one robot is remaining then relocation of a pair of consecutive robots jointly is chosen (3-5) otherwise the last robot is placed as single. The placement of two consecutive robots – say robots $r$ and $s$ – first decides if it is better to first move $s$ next to $r$ (line 14) or vice versa (line 20) or not use the joint relocation at all (line 13). Not using the joint relocation can be used as the third option. The decision in the presented pseudo-code is based upon lengths of shortest paths between current positions of robots; however, it can be done according to other criteria. We used simulation to calculate the number of moves necessary for each of the two options in the actual C++ implementation used in experiments (this is the major reason why snake enhanced *PUSH-and-SWAP* consumes order of magnitude more time than the original version). The option that consumes fewer moves is finally chosen.

**Algorithm 4.** *The* ***simplified PUSH-and-SWAP with snakes*** *algorithm.* The algorithm solves cooperative path-finding problem (CPF) over arbitrary graphs with at least three unoccupied vertices. It is a simplified version of the original *PUSH-and-SWAP*; the simplification consists in omitting treatments of special cases that occur with general graphs. Hence, the presented algorithm is incomplete for arbitrary graphs. The simplification however does not affect the behavior of the snake improvement over tested puzzle and random bi-connected graph instances.

- Lock($U$) — locks all the vertices from set $U$; each vertex is either *locked* or *unlocked*; a robot can be moved out of the locked vertex if it is later returned back (a case of *Swap* and *Twin-Swap* operations)
- Unlock($U$) — unlocks all the vertices from set $U$
- Move-Robot($r, v$) — moves robot $r$ from its current location to vertex $v$
- Make-Unoccupied($v$) — vacates vertex $v$ sensitively to locked vertices
- Make-Unoccupied($r, s, t, v$) — vacates vertex $v$ irrespective to locked vertices but preserving positions of robots $r$, $s$, and $t$
- Relocate-Robot($r, v$) — relocates robot $r$ from its current location to $v$ and locks it there; this function is implemented within the original algorithm – it uses *Push* and *Swap* basic operations
- Start-Undo() — start recording of performed moves for later undoing them (execute reverse moves in the reverse order); saves the current robot arrangement
- Stop-Undo() — stops recording of performed
- Execute-Undo() — undoes the sequence of recorded moves; clears the recorded undo sequence
- Cancel-Undo() — clears the recorded undo sequence and restores robot arrangement saved at the beginning of undo recording
- Twin-Exchange($r, s, t, v$) — exchanges a triple of robots $r$, $s$, and $t$ at vertex $v$ having at least 4 neighbors
- Make-Unoccupied-3($r, s, t, v$) — vacates 3 neighbors of vertex $v$ while positions of robots $r, s, t$ with $t$ standing at $v$ is preserved

**function** *PUSH-and-SWAP-Snake-Solve*($G = (V, E), R, \alpha_0, \alpha^+$)**: boolean**

/* Top level function of the *PUSH-and-SWAP* algorithm with snakes;
solves a given cooperative path-finding problem.
Returns $TRUE$ if succeeds or $FALSE$ in case of failure (untreated case).
Parameters: $G$ - a graph modeling the environment,
$R$ - a set of robots,
$\alpha_0$ - a initial configuration of robots,
$\alpha_+$ - a goal configuration of robots. */

**let** $R = [r_1, r_2, \ldots, r_\mu]$ some ordering of robots
**for** $i = 1,2, \ldots, \mu$ **do**
  **if** $i < \mu - 1$ **then**
    **if not** *Relocate-Twin-Robots*($r_i, r_{i+1}, \alpha_+(r_i)$) **then**
      **return** $FALSE$
  **else**
    **if not** *Relocate-Robot* ($r_i$) **then**
      **return** $FALSE$
**return** $TRUE$

**modification** w.r.t. original *PUSH-and-SWAP*

```
function Relocate-Twin-Robots(r, s, v): boolean
    /* Relocates a pair of robots jointly towards the goal vertex of the first of them.
    Returns TRUE if succeeds or FALSE in case of failure (no vertex of degree at least 4).
    Parameters:   r, s - a pair of robots,
                  v – goal vertex of r. */
let u ∈ V be a neighbor of α(r)
let w ∈ V be a neighbor of α(s)
d_tandem ← min{dist_G(α(s), u) + dist_G(α(r), v), dist_G(α(r), w) + dist_G(α(s), v)}
d_separate ← min{dist_G(α(r), α⁺(r)) + dist_G(α(s), α⁺(s))}
if d_tandem < d_separate then
    if dist_G(α(s), u) + dist_G(α(r), v) < dist_G(α(r), w) + dist_G(α(s), v) then
        /* It seems to be better to move s next to r and then move together towards v. */
        if Relocate-Robot(s, u) then
            let π = [u = p_1, α(r) = p_2, p_3, ..., p_k = v] be the shortest
                path connecting a positions of r and s with v
        else
            return FALSE
    else
        /* It seems to be better to move r next to s and then move together towards v. */
        if Relocate-Robot(r, w) then
            let π = [α(s) = p_1, w = p_2, p_3, ..., p_k = v] be the shortest
                path connecting a positions of r and s with v
        else
            return FALSE
    for i = 1,2, ..., k − 2 do
        if not Twin-Push(p_i, p_{i+1}, p_{i+2}) then
            if not Twin-Swap(p_i, p_{i+1}, p_{i+2}) then
                return FALSE
    if not Relocate-Robot(s, α⁺(s)) then
        return FALSE
else
    if not Relocate-Robot(r, α⁺(r)) then
        return FALSE
    if not Relocate-Robot(s, α⁺(s)) then
        return FALSE
return TRUE
```

```
function Twin-Push(r, s, v): boolean
if α⁻¹(v) ≠⊥ then
    Lock({α(r), α(s)})
    if not Make-Unoccupied(v) then
        Unlock({α(r), α(s)})
        return FALSE
Unlock({α(r), α(s)})
Move-Robot(r, v)
Move-Robot(s, u)
```

**function** *Twin-Swap*$(r, s, v)$**: boolean**
46: $T \leftarrow \{u \in V | \deg_G(u) \geq 4\}$
47: **for each** $u \in T$ **do**
48: *Start-Undo*()
49: $t \leftarrow \alpha^{-1}(v)$
50: **if** *Twin-Multipush*$(r, s, t, u)$ **then**
51: **if not** *Make-Unoccupied-3*$(r, s, t, u)$ **then**
52: *Cancel-Undo*()
53: **continue**
54: *Stop-Undo*()
55: *Twin-Exchange*$(r, s, t, u)$
56: *Execute-Undo*()
57: **return** $TRUE$
58: **else**
59: *Cancel-Undo*()
60: **return** $FALSE$

---

**function** *Twin-Multipush*$(r, s, t, v)$**: boolean**
61: **let** $\pi = [\alpha(r) = p_1, \alpha(s) = p_2, \alpha(t) = p_3, \dots, p_k = v]$ be the shortest
62: path connecting a positions of $r$, $s$, and $t$ with $u$
63: **for each** $i = 1,2, \dots, k-3$ **do**
64: **if not** *Make-Unoccupied*$(r, s, t, p_{i+3})$ **then**
65: **return** $FALSE$
66: *Move-Robot*$(t, p_{i+3})$
67: *Move-Robot*$(r, p_{i+2})$
68: *Move-Robot*$(s, p_{i+1})$
69: **return** $TRUE$

---

Having the consecutive robots $r$ and $s$ next to each other they are moved jointly towards the goal of the former robot $r$ (lines 24-27). If robots are ordered topologically according to their goals – that is, robots that are close to each other in the ordering have their goal close to each other in the graph – the later robot $s$ should appear close to its goal after the relocation of the pair. If the later robot is not in its goal after the joint relocation, the final single robot relocation corrects this (lines 28-29). It is optimistically assumed that final single robot relocation does not produce too many moves thanks to chosen ordering of robots.

The joint relocation of the pair relies on *Twin-Push* and *Twin-Swap* operations. After finding a path that connects the current occurrence of the pair of robots with the goal of the former one, the pair of robots is moved along this path. It is assumed that robot $r$ is the leader of the snake followed by robot $s$.

The *Twin-Push* operation is applicable if a vertex on the path in front of the leading robot $r$ can be freed by moving robots in not yet locked vertices (line 40). After vacating the vertex in front of $r$, the pair is moved one step forward. If the operation *Twin-Push* fails, that is, if the next vertex on the path is occupied and cannot be freed by moving robots in unlocked vertices, then *Twin-Swap* operation is applied.

Assume that robot $t$ resides in the next vertex on the path so it is the task to jump with $r$ and $s$ over $t$. This can be done by finding a vertex with at least 4 neighbors $v$ to which the triple is moved by the *Twin-Multipush* auxiliary operation (line 50) and where the order of the three robot is changed from $[t, r, s]$ to $[r, s, t]$ (the change of the order is illustrated in Figure 11). The relocation of the triple of robots by *Twin-Multipush* to $v$ and subsequent freeing of the neighbors of $v$ by *Make-Unoccupied-3* (line 44) to enable order change disregards vertex locking.

Hence, robots in locked vertices may move out of their goals. They are moved back to their goals by undoing all the moves (supported by auxiliary operations *Start-Undo*, *Stop-Undo*, *Cancel-Undo*, and *Execute-Undo*) that relocated the triple of robots to $v$ and vacated its neighbors (moves that caused the change of the ordering are not undone). Undoing the moves (lines 48-56) preceding the change of ordering of the triple results in the situation, in which order of the three robots is changed at their current position while all other robots reside in their positions as well.

Regarding the ordering of vertices of degree at least 4 to make the *Twin-Swap possi*ble, the nearest such vertices to the triple being exchanged are tried first in our implementation. The same implementation is used within the *Swap* operation where a vertex of the degree at least 3 is being searched.